\documentclass[letterpaper,12pt]{article}
\pdfoutput=1
\usepackage{lscape,enumitem}
\usepackage{arabtex}
\usepackage{rotating}
\newcommand\independent{\protect\mathpalette{\protect\independenT}{\perp}}
\def\independenT#1#2{\mathrel{\rlap{$#1#2$}\mkern2mu{#1#2}}}
\usepackage{graphicx,textcomp}
\usepackage{natbib}
\usepackage{vmargin}
\setpapersize{USletter}
\usepackage{color}
\usepackage[reqno]{amsmath}
\usepackage{amsthm}
\usepackage{framed}
\usepackage{amssymb,enumerate}
\usepackage[all]{xy}
\usepackage{lscape}
\usepackage{tikz}
\usepackage{tabularx}
\usetikzlibrary{fit,positioning}
\usetikzlibrary{arrows}

\newtheorem{ass}{Assumption}
 \numberwithin{equation}{section}
 \numberwithin{equation}{section}
\usepackage{dcolumn}
\newcolumntype{.}{D{.}{.}{-1}}
\newcolumntype{d}[1]{D{.}{.}{#1}}
\usepackage{url}

\usepackage{color,setspace}
\definecolor{spot}{rgb}{0.6,0,0}
\definecolor{myred}{rgb}{.803, .36,.36}
\definecolor{myblue}{rgb}{0,0.1,0.6}
\definecolor{mygreen}{rgb}{0.1,0.8,0.2}

\title{How to Make Causal Inferences Using Texts\thanks{We thank Edo Airoldi, Peter Aronow, Matt Blackwell, Sarah Bouchat, Chris Felton, Mark Handcock, Erin Hartman, Rebecca Johnson, Gary King, Ian Lundberg, Rich Nielsen, Thomas Richardson, Matt Salganik, Melissa Sands, Fredrik S{\"a}vje, Arthur Spirling, Alex Tahk, Endre Tvinnereim, Hannah Waight, Hanna Wallach, Simone Zhang and numerous seminar participants for useful discussions about making causal inference with texts.  We also thank Dustin Tingley for early conversations about potential SUTVA concerns with respect to STM and sequential experiments as a possible way to combat it.}}
\author{Naoki Egami\thanks{Ph.D. Candidate, Department of Politics, Princeton University, negami@princeton.edu} \and Christian J. Fong\thanks{Ph.D. Candidate, Graduate School of Business, Stanford University, cjfong@stanford.edu} \and Justin Grimmer\thanks{
    Associate Professor, Department of Political Science, University of Chicago,
    JustinGrimmer.org, grimmer@uchicago.edu.} \and Margaret E. Roberts\thanks{Assistant Professor, Department of Political Science, University of California San Diego, meroberts@ucsd.edu} \and Brandon M. Stewart\thanks{Assistant Professor, Department of Sociology, Princeton University, brandonstewart.org, bms4@princeton.edu}}
\date{\today}

\begin{document}
\maketitle
\begin{abstract}
\noindent New text as data techniques offer a great promise: the ability to inductively discover measures that are useful for testing social science theories of interest from large collections of text.  We introduce a conceptual framework for making causal inferences with discovered measures as a treatment or outcome.  Our framework enables researchers to discover high-dimensional textual interventions and estimate the ways that observed treatments affect text-based outcomes. We argue that nearly all text-based causal inferences depend upon a latent representation of the text and we provide a framework to learn the latent representation. But estimating this latent representation, we show, creates new risks: we may introduce an identification problem or overfit.  To address these risks, we describe a split-sample framework and apply it to estimate causal effects from an experiment on immigration attitudes and a study on bureaucratic response.  Our work provides a rigorous foundation for text-based causal inferences.
\end{abstract}

\newpage


\newpage 

\singlespacing
\section{Introduction}

One of the most exciting aspects of research in the digital age is the rapidly expanding evidence base for social scientists-- from judicial opinions to political propaganda, Twitter, and government documents \citep{king2009changing, salganik2017bit, grimmer2013text}.  Text is now regularly combined with new computational tools to measure quantities of interest.  This includes applications of hand coding and supervised methods that assign texts into predetermined categories \citep{boydstun2013making}, clustering and topic models that discover an organization of texts and then assigns documents to those categories \citep{catalinac2016pork}, and factor analysis and item-response theory models that embed texts into a low-dimensional space \citep{spirling2012us}.  Reflecting the proliferation of data and tools, scholars increasingly use text-based methods as either the dependent variable or independent variable in their studies.  Yet, in spite of the widespread application of text-based measures in causal inferences and a flurry of exciting new social science insights, the existing scholarship often leaves unstated the assumptions necessary to identify text-based causal effects.  

In this paper we provide a conceptual framework for text-based causal inferences, building a foundation for research designs using text as the outcome or intervention.  Our paper connects the text as data literature \citep{lasswell1938propaganda, laver2003extracting, pennebaker2003psychological, quinn2010analyze}, 
with the growing literature on causal inference in the social sciences \citep{pearl2009causality, imbens2015causal, hernan2018causal}.
The key to connecting the two traditions is recognizing the central role of \emph{discovery} when using text data for causal inferences.  

Discovery is central to text-based causal inferences because text is complex and high-dimensional and therefore requires simplification before it can be used for social science.  This simplification can be intuitive and familiar.  For example, we might take a collection of emails and divide them into `spam' and `not spam.' We call the function which maps the documents into our measure of interest $g$.  We think of $g$ as a \emph{codebook} that tells us how to compress our documents into categories, topics, or dimensions.  $g$ plays a central role in causal inference using text.

The need to discover and iteratively define measures and concepts from data is a fundamental component of social science research \citep{tukey1980we}.  One of the most compelling promises of modern text analysis is the capacity to help researchers discover new research questions and measures inductively.  However, the iterative discovery process poses problems for causal inference. We may not know $g$ in advance of conducting our experiment and consequently, we may not know our outcome or treatment. We describe an identification and estimation problem that arises from a common source --- using the same documents for discovery of measures and the estimation of causal effects. To resolve both problems we introduce a procedure and a set of sufficient assumptions for using text data in research designs. 

The identification problem occurs because the particular $g$ we obtain will often depend upon the treatments and responses, and using this information can create a dependence across units.  Most causal inference approaches assume that each unit's response depends on only its treatment status and not any other unit's treatment.  This is one component of the Stable Unit Treatment Value Assumption (SUTVA) \citep{rubin1980comment}.  But, when using the same documents for discovering $g$ and estimating effects the analyst can induce a SUTVA violation where none had previously existed.  This arises because the $g$ that we discover depends on the particular set of treatment assignments and responses in our sample, so that changing other units' treatment status will change the $g$ discovered and, as a result, the measured response or intervention for a particular unit.  We call this dependence an \textit{Analyst Induced SUTVA Violation} (AISV) because the analyst induces the problem when estimating $g$ even in an experiment where there is otherwise no dependence across units.  The AISV problems are substantial: if an AISV occurs it makes it impossible to evaluate properties of our estimator such as variance, bias or consistency without further assumptions.

Even if we dismiss or assume away the identification problem, the complexity of text leads to an estimation problem: \textit{overfitting}.  By using the same documents to discover and estimate effects, even well-intentioned analysts may mistake noise for a robust causal effect.  The dangers of searching over $g$ is a more general version of the problem of researchers recoding variables in an experiment to search for significance.  This idea of overfitting also formalizes the intuition that some analysts have that latent-variable models are `baking in' an effect.

In this paper, we introduce a procedure to diagnose and address both problems in service of our ultimate goal---finding replicable and theoretically relevant causal effects.  
We adopt a solution which simulates a fresh experiment: a train/test split (also called a split-sample design).  
While a train/test split is used regularly to assess the performance of classifiers, it has only more recently been used to improve causal inference \citep{wager2017estimation, fafchamps2017using, chernozhukov2017double}. We show how a train/test split avoids the problems text data present for causal inference.  This connects to the general principle of separating the specification of potential outcomes from analysis \citep{imbens2015causal}.  Splitting our sample separates a training set for use in discovery (fixing potential outcomes) from a test set for use in estimation (analysis), conditional on the discovered $g$.  The estimate in the test set provides insight into what the results from a new experiment would be and, as we show below, resolves our identification and estimation problems.  Splitting the sample, then, enables discovering $g$ while facilitating causal effects.

\begin{figure}
\caption{Our Procedure for Text-Based Causal Inferences}\label{f:procedure}
\begin{center}
\includegraphics[width=.6\textwidth]{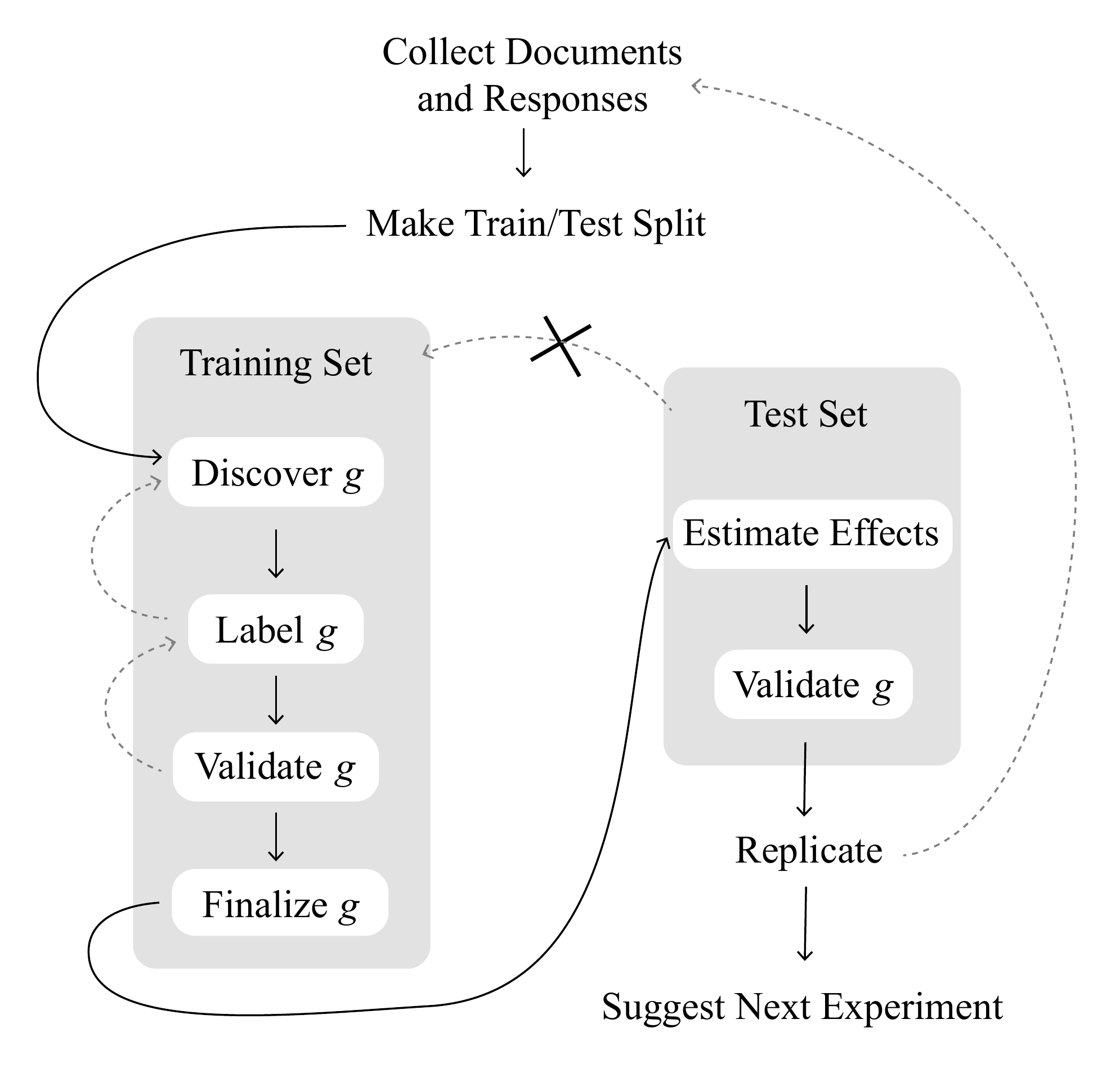}
\end{center}
\end{figure}

Building on the split sample approach to obtain and apply $g$, we explain our suggested procedure (Figure~\ref{f:procedure}) and then apply it to two specific examples: text as the dependent variable and the text as the independent variable.  In Appendix \ref{app:procedure} we provide a verbal description of Figure~\ref{f:procedure}.   

To introduce this procedure our paper proceeds as follows. In Section~\ref{sec:central} we provide a definition of $g$ and describe the central role that it plays in text analysis.  Section~\ref{sec:problem} discusses the core identification and estimation concerns that complicate the use of $g$ in a causal inference setting. Having described the problem, in Section~\ref{sec:solution} explains why sample splitting is a solution, describing how it works, how it addresses the core problems we raise, and the trade-offs in its use.  We also defer discussion of prior work until this section so that we can show how our work connects to a long-tradition of sample-splitting approaches in machine learning and more recently in causal inference.  In Section~\ref{sec:application}, we illustrate our approach using applications in two settings: text as outcome and text as treatment.  Section~\ref{sec:conclusion} concludes and an online appendix provides additional technical details, proofs of key claims, and details on statistical methods used in our applications.

\section{The Central Role of $g$, The Codebook Function}
\label{sec:central}

The central problems that we address stem from the need to compress text data to facilitate causal inference.  The codebook function, $g$, compresses high-dimensional text to a low-dimensional measure used for the treatment or outcome.  In this section we explain why $g$ is essential, how to obtain $g$, and how to evaluate candidate $g$'s.

\subsection{What is $g$ and why do we need it?}

The codebook function, $g$, is essential because the text is typically not usable for social science inference in its \textit{raw} form.  Social scientists are often interested in some emergent property of the text---such as the topic that is discussed, the sentiment expressed, or the ideological position taken. Documents are high-dimensional, complicated, and sparse. The result is that distinct blocks of text can convey similar topics or sentiment.  Reducing the dimensions of the text allows us to group texts and make inferences from our data.

Suppose we are interested in understanding how candidate biographies influence the popularity of a candidate.   Each biography is unique, so we cannot estimate the effect of any individual biography on a candidate's popularity.  Instead, we are interested in some latent property of the text's effect on the popularity of the candidate, such as occupational background.  In this example $g$ might compress the text of the biography into an indicator of whether the candidate is a lawyer.  The analyst could define $g$ in numerous ways including hand-coding.  $g$ could also be defined automatically from the text, by looking for the presence or absence of the word ``lawyer", or a group of words or phrases that convey that someone has a legal background, such as ``JD", ``attorney", and ``law school".  Being a lawyer is just one latent feature in the text.  Different $g$'s might measure if a candidate held prior office, went to college, or served in the military.  

Our most consequential decision about $g$ is the space we compress the text into.  Options for this space could include discrete categories, proportions, or continuous variables (like ideal point estimates). We will call the lower-dimensional space $\mathcal{Z}$.  Typically these low-dimensional representations are then given a label for interpretation. For example, we might use $g$ to bin social media posts into ``positive,'' ``negative,'' or ``neutral,'' or,  put portions of documents into topics that we label ``Sports,'' ``Weather,'' or ``Politics.''

Social scientists working on text as data have adopted this compression approach, although the low-dimensional representation is often only implicit \citep{laver2003extracting, grimmer2012words, spirling2012us, catalinac2016pork}. We can also think of $g$ as the \textit{codebook function} because it plays the role of a codebook in a manual content analysis, describing a procedure for \textit{organizing} the researcher's texts in some systematic way. $g$ takes on a central role because it connects the raw text to the underlying property that the researcher cares about.  While applied work on measurement often describes the categories under study, discussion of the implications of $g$ as an object of interest is rare. Nevertheless, $g$ is always implicitly present in any systematic analysis of text---any instance where a set of documents is placed into a common set of categories or is assigned a common set of properties. Once a researcher decides on and estimates $g$, then text is usually ready to be used in statistical analysis.

\subsection{Discovering $g$}
\label{sub:discoveringG}

While $g$ is necessary to make causal inference, rarely is it determined exactly from a theory or prior research.  Even in manual content analysis \citep{krippendorff2004content,neuendorf2016content}, researchers typically read at least a portion of the documents to write a codebook that determines how coders should put documents into the categories of interest.  More recently, a wide array of machine learning methods are used to discover $g$ from the data \citep{blei2003latent, hopkins2010method}.  These newly discovered categories can help shape research questions, identify powerful textual interventions, and capture text-based outcomes.

In spite of its central role across forms of text analysis, social scientists rarely discuss the process of discovery that lead to a particular codebook. In practice, these coding schemes are developed through iteration between coding rules and the documents to be coded.  We raise two main points about the discovery of $g$ that apply regardless of the methodology applied.

\paragraph{1) We can (and often do) learn $g$ from the data.} There are three strategies for learning $g$ from the data.  First, we could read a sample of text.  In manual content analysis, $g$ often relies on some familiarity with the text or reading a sample of documents to decide how the text should map into categories.  Second, we could use a method to classify texts into categories using hand coded examples for training.  Supervised methods, which are conceptually similar to manual content analysis, use statistical and algorithmic methods attempting to estimate the best $g$ from hand coded or otherwise labeled documents.  Last, unsupervised learning discovers a low-dimensional representation and assigns documents to that representation.

\paragraph{2) There is no single correct $g$.} Regardless of the methods used in discovery, the analyst chooses a $g$ on the basis of their theoretical question of interest. Different theories imply different organizations of the text and, therefore, different $g$'s. However, we can and \textit{should} evaluate $g$ once we have defined a question of interest.  Given a particular function and a particular purpose, we can label the identified latent features, the scales measured, and the classification accuracy.  The \emph{post hoc} validation of $g$ provides clarity for both the researcher and the reader to correctly interpret the underlying latent features \citep{grimmer2013text}.  Our goal in the validation is to ensure that the interpretation implicit in our theoretical argument arises from and corresponds with the mapping in our chosen $g$.

\subsection{Finalizing $g$}
\label{sub:selectingG}

Although there is no application-independent correct $g$, once we have a question of interest, there are properties of $g$ that are useful:  interpretability, theoretical interest, label fidelity, and tractability.

\paragraph{Property 1: Interpretability} First, $g$ should be \textit{interpretable}.  To claim that a measure is theoretically interesting, we have to interpret it.  Interpretability is research and text specific, but our articles must communicate to the reader what the measure in a specific study is capturing.  This is particularly important for $g$'s discovered from text data, which are based on underlying covariances in the data and thus will not necessarily be interpretable.

\paragraph{Property 2: Theoretical Interest} The codebook function should also create measures of \textit{theoretical interest}.  We want to find low-dimensional representations of text that operationalize concepts from a theory and identify causal effects that test observable implications of the theory.  
Ideally, we would like to focus on large magnitude causal effects.  All else equal, larger effects help us to explain more of the behavior of theoretical interest.      

\paragraph{Property 3: Fidelity} 
We also want to choose functions with high \textit{fidelity} between the label we give to the components of $g$ and the text it is compressing.  Establishing fidelity involves producing evidence that the latent variable $\boldsymbol{z}$ accurately captures the property implied by the label. This is a common exercise in the social sciences; there is always an implicit mapping between the labels we use for our variables and the reality of what our labels measure.  For text analysis, we think of maximizing label fidelity as minimizing the surprise that a reader would have in going from the label to reading the text. Fidelity is closely connected to the literature on validity in measurement and manual content analysis \citep[see e.g.,][]{grimmer2013text, quinn2010analyze, krippendorff2004content}.

\paragraph{Property 4: Tractable} Finally, we want the development and deployment of $g$ to be \textit{tractable}.  In the context of manual content analysis this means the codebook can be applied accurately and reliably by human coders and that the number of documents to be coded is feasible for the resources available.  In the case of learning $g$ statistically, tractability implies that we have a model which can be estimated using reasonable computational resources and that it is able to learn a useful representation with the number of documents we possess.

\bigskip  

There is an inherent tension between the four properties.  This is most acute with the tension between theoretical interest and label fidelity.  It is often tempting to assign a very general label even though $g$ is more specific.  This increases theoretical relevance, but lowers fidelity.  The consequence can be research that is more difficult to replicate. Alternatively, we might have a $g$ that coincides with a label because it increases the chances that our result can be replicated.  But this could reduce the theoretical interest.  

The analog of $g$ lurks in every research design, including those that use standard data.  Invariably when making an argument the researcher needs to find empirical surrogates or operationalized the concepts in her theoretical argument.  For example, every time a researcher uses gross domestic product (GDP) as a stand-in for the size of the economy, she is projecting a high-dimensional and complicated phenomenon---the economy---into a lower-dimensional and more tractable variable---GDP.  The causal estimand is defined in terms of its effect on GDP, but the theoretical argument is made about the size of the economy. While there is no correct measure to use for the economy, the reader can and should still interrogate the degree to which the chosen measure appropriately captures the broader theoretical concept that the researcher wants to speak to.

\section{The Problem of Causal Inference with $g$}
\label{sec:problem}
Text is high-dimensional, so we use the codebook function, $g$, to learn a low-dimensional representation to make inferences.  But using $g$ to compress text introduces new problems for causal inference.  In this section we explain how $g$ facilitates causal inference with text and then characterize the problems it creates.\footnote{At a technical level we can think of an experiment with the process of discovery as a form of data-adaptive estimation \citep{van2013statistical}, a framework which originates from biostatistics and describes circumstances where our target estimation is not fixed in advance.}  In Section \ref{sub:notation} we place $g$ in the traditional causal inference setting.  Section \ref{sub:the_problem} explains how the use of $g$ leads to the problems of an \emph{analyst induced SUTVA violation} and \emph{overfitting}.     

\subsection{Causal inference with $g$}
\label{sub:notation}

To begin, we review potential outcomes notation and assumptions used when there is no text or dimensionality reduction and we are analyzing a unidimensional treatment and outcome \citep{imbens2015causal}.  Denote our dependent variable for each unit $i$ ($i \in 1,2,\dots,N)$ with $Y_{i}$, the treatment condition for unit $i$ will be $T_{i}$.  We define the space of all possible outcomes as $\boldsymbol{\mathcal{Y}}$ and the space of all possible treatments as $\boldsymbol{\mathcal{T}}$. When the treatment is binary we refer to $Y_i(1)$ as the potential outcome for unit $i$ under treatment and $Y_i(0)$ as the potential outcome under control and the individual causal effect (ICE) for unit $i$ is given by $\text{ICE}_i = Y_i(1) - Y_i(0)$.  Our typical estimand is some function of the individual causal effects such as the average treatment effect (ATE), $E[Y_i(1) - Y_i(0)]$.   

To identify the average treatment effect using a randomized experiment we make three key assumptions.  First, we assume that the response depends only on the assigned treatment, often called the Stable Unit Treatment Value Assumption (SUTVA).  Specifically:
\begin{ass}[SUTVA] \label{a:SUTVA}
For all individuals $i$, $Y_{i}(T)  = Y_{i}(T_i).$
\end{ass}
\noindent Second, we will assume that our treatment is randomly assigned:
\begin{ass}[Ignorability]\label{a:random}
$Y_{i}(t)  \independent  T_i$
\end{ass}
\noindent Third,we will assume that every treatment has a chance of being seen:
\begin{ass}[Positivity]\label{a:positivity}
$Pr(T_i=t) > 0 $ for $t \in \mathcal{T}.$
\end{ass}

\noindent The second and third assumptions are guaranteed by proper randomization of the experiment whereas the first is an assumption that is generally understood to mean that there is no interference between units and no hidden values of treatment.  For each observation we observe only a single potential outcome corresponding to the realized treatment.

Building off of this notation, we can introduce mathematical notation to cover high-dimensional text and the low-dimensional representation of texts derived from $g$ that we will use for our inferences.  We start by extending our notation to cover multi-dimensional outcomes, $\boldsymbol{Y}_{i}$, and multi-dimensional treatments, $\boldsymbol{T}_i$.  We will suppose, for now, that we have already determined $g$, the codebook function. Recall $g$ is applicable regardless of whether the coding is done by a machine learning algorithm, a team of undergraduate research assistants or an expert with decades of experience.

We write the set of possible values for the mapped text as $\mathcal{Z}$ with a subscript to indicate if it is the dependent variable or treatment.  We denote the realized values of the low-dimensional representation for unit $i$ as $\boldsymbol{z}_i$ ($i=1, \hdots, N)$.  We suppose that when the outcome is text $g:\mathcal{Y} \rightarrow \mathcal{Z}_Y$ and $g(\boldsymbol{Y}_i) = \boldsymbol{z}_i$, and when the treatment is text $g:\mathcal{T} \rightarrow \mathcal{Z}_T$ and  $g(\boldsymbol{T}_i) = \boldsymbol{z}_i$. The set $\mathcal{Z}$ is a lower-dimensional representation of the text and can take on a variety of forms depending upon the study of interest.  For example, if we are hand coding our documents into two mutually-exclusive and exhaustive categories, then $\mathcal{Z}$ is $\{0,1\}$.  If we are using a mixed-membership topic model to measure the prevalence of $K$ topics as our dependent variable, then $\mathcal{Z}$ is a $K-1$ dimensional simplex.  And if we are using texts as a treatment, we might suppose that $\mathcal{Z}$ is the set of $K$ binary feature vectors, representing the presence or absence of an underlying treatment (see Appendix~\ref{app:binary} for the reason we prefer binary treatments, though continuous treatments also fit within our framework).  
There are numerous other types of $g$ that we might use---including latent scales, dictionary-based counts of terms, or crowd-sourced measures of content. 
The only requirement for $g$ is that it is a function.     

We next use $g$ to write our causal quantity of interest in terms of the low-dimensional representation.  To make this concrete, consider a case where we have a binary non-text treatment and a text-based outcome (we consider other causal estimands below).  Suppose we hand code each document into one of $K$ categories such that for unit $i$ we can write the coded text under treatment as $g(\boldsymbol{Y}_i(1)) = \boldsymbol{z_i}(1)$.  We can then define the average treatment effect for category $k$ to be: 
\begin{eqnarray}
\text{ATE}_{k} & = &   E[g(\boldsymbol{Y}_{i}(1))_{k}  - g(\boldsymbol{Y}_{i}(0))_{k} ] \label{e:ATE} \\
               & = &   E[z_{i, k}(1)  - z_{i,k}(0) ] \nonumber
\end{eqnarray}
where $z_{i,k}(1)$ indicates the value of the $k$-th category, for unit $i$, under treatment.

\subsection{The Problems: Identification and Overfitting}
\label{sub:the_problem}

Equation \ref{e:ATE} supposes that we already have a $g$ in hand.  As we mentioned above, $g$ is often discovered by interacting with some of the data, either by reading or through machine learning.  To describe this problem more clearly, we denote the set of documents considered in development of $g$ as $\boldsymbol{J}$ and write $g_{\boldsymbol{J}}$ to indicate the dependence of $g$ on the documents.  Problems of identification and estimation arise where the set of documents used to develop $g$, $\boldsymbol{J}$, overlaps with the set of documents used in estimation which we will call $\boldsymbol{I}$.  There are two broad concerns: an identification problem arising from an \textit{Analyst Induced SUTVA Violation} (AISV) and an estimation problem with overfitting.  

\subsubsection{Identification concerns: Analyst Induced SUTVA Violations}
\label{sub:identification}

If Assumption 1 holds then each observation's response does not depend on other units' treatment status.  But even when Assumption 1 holds, when we discover $g_{\boldsymbol{J}}$, we can create a dependence across observations in $\boldsymbol{J}$ because the particular randomization may affect the $g_{\boldsymbol{J}}$ we estimate.  This violation occurs because the treatment vector $\boldsymbol{T}_{\boldsymbol{J}}$ -- the treatment assignments for all documents $\boldsymbol{J}$-- affects the $g$ that we obtain, inducing dependence across \textit{all} observations in $\boldsymbol{J}$.  If we then try to use the documents in $\boldsymbol{J}$ for estimation of the effect, we have violated SUTVA.  This violation is induced by the analyst in the process of discovering $g$, which is why we call it an \emph{Analyst} induced violation.  Appendix~\ref{app:technical} provides a formal definition of AISV.

To see how the AISV works in practice, consider a stylized experiment on four units with a dichotomous intervention (treatment/control) and a text-based outcome.  We might imagine potential outcomes that have a simple relationship between treatment and the text-based outcome such as the one shown in Table~\ref{tab:stylized}.  Treated units talk about Candidate Morals and Polarization and control units talk about Taxes and Immigration.

\begin{table}[ht!]
\begin{center}
\begin{tabular}{r|cc}
& Treated	& Control \\
\hline
Person 1 &	Candidate Morals &	Taxes \\
Person 2 &	Candidate Morals &	 Taxes \\
Person 3 &	Polarization &	Immigration \\
Person 4 &	Polarization &  Immigration
\end{tabular}
\end{center}
\caption{A stylized experiment indicating the potential outcomes of a textual response.\label{tab:stylized}}
\end{table}

Using Table \ref{tab:stylized} we can imagine the properties of an estimator applied to this text-based experiment as we re randomize.  Suppose that for each randomization we decide on both the form of $g$ and estimate the treatment effect given $g$.  For example, consider if we observe the treatment vector (1,1,0,0), we would observe only two of the four categories: morals and immigration.  A reasonable $g$ might compress the text based responses to two variables: an indicator variable for discussing morals and an indicator variable for discussing immigration.  If we randomize again and then we get (1,0, 1, 0) we observe all four categories.  In this case, $g$ might map the text based responses to a four-element long vector, with an indicator for whether each distinct category is discussed in the response.  Under a third randomization (0,0,1,1) we are back to only two categories: taxes and polarization; so $g$ might be two bivariate indicator variables, with the categories corresponding to whether someone discussed taxes or not or polarization or not.  

As we randomize we estimate new $g$'s with different categories. This lack of \emph{category stability} complicates our ability to analyze our estimators as we traditionally do, using a framework based on re-randomization.  We take this category and classification stability for granted in standard experiments because categories are defined and fixed before the experiment.  But when we estimate categories from data the discovered $g$ depends on the randomization and therefore dependence between units is induced by the analyst. And even if we fix the categories, as we might do with a supervised model, different randomizations may lead to different rules for assigning documents to categories, leading to a lack of \emph{classification stability}. If, however, we fix $g$ before estimating the effects, the problem is solved.

\subsubsection{Estimation concerns: Overfitting}
\label{sub:estimation}

Even if we assume away the AISV, estimating $g$ means that researchers might \emph{overfit}: discover effects that are present in a particular sample but not in the population. This is a particular risk when researchers are searching over different $g$'s to find those that best meet the criteria of interpretability, interest, fidelity and tractability.  The overfitting problem is particularly acute when a researcher is fishing --- searching over $g$'s to obtain statistical significance or estimates that satisfy a related criterion.  But overfitting can occur even if researchers are conducting data analysis without ill-intentions. This happens because following best practice with almost all available text as data methods requires some iteration.  With hand coding iteration occurs to refine the codebook, with supervised models it occurs when we refine a classifier, and with unsupervised methods it happens as we adjust parameters to examine new organizations. 

Fishing and overfitting are a problem in all experimental designs and not just those with text. The problem of respecifying $g$ until finding a significant result is analogous to the problem of researchers recoding variables or ignoring conditions in an experiment, which can lead to false-positive results. \citep{simmons2011false}.
The problem with text-based inferences is heightened because texts are much more flexible than other types of variables, creating a much wider range of potential $g$'s.  This wider range increases the risk of overfitting, even amongst well-intentioned analysts.  Overfitting is also likely in texts because it is so easy to justify a particular $g$ after the fact -- the human brain is well-equipped to identify and justify a pattern in a low-dimensional representation of text, even if that pattern emerges merely out of randomness.  This means that validation steps alone may be insufficient safeguard against overfitting, even though texts provide a rich set of material to validate the content.

\section{A Train/Test Split Procedure for Valid Causal Inference with Text}
\label{sec:solution}

To address the identification issues caused by the AISV and the estimation challenges of overfitting, we must break the dependence between the discovery of $g$ and the estimation of the causal effect.  The most straightforward approach is to define $g$ before looking at the documents.  Defining the categories beforehand, however, limits our coding scheme, excluding information about the language used in the experiment's interventions or what units said in response to a treatment.  If we define our codebook before seeing text we will miss important concepts and have a poorer measure of key theoretical concepts.

We could also assume the problem away.  Specifically, to eliminate the AISV it is sufficient to assume that the codebook that we obtain is invariant to randomization.  Take for example the text as outcome case; if over different randomizations of the treatment the $g$ we learned does not change, we don't have an AISV. We define a formal version of this assumption in Appendix~\ref{app:away}.  

Our preferred procedure is to explicitly separate the creation of $g$ and the estimation of treatment effects.  This procedure avoids the AISV and provides a natural check against overfitting.  To explicitly separate the creation of the codebook and its application to estimate effects, we randomly divide our data into a training set and a test set.  
Specifically, we randomly create a set of units in a training set denoted by the indices $\boldsymbol{J}$ and a non-overlapping test set denoted by the indices $\boldsymbol{I}$.  
We use only the training set to estimate the $g_{\boldsymbol{J}}$ function and then discard it.  We then use the test set exclusively to estimate the causal effect on the documents in $\boldsymbol{I}$.

This division between the training and test set addresses both the identification and estimation problems.  It avoids the AISV in the test set because the function $g$ does not depend on the randomization in the test set, so that each test set unit's response depends only on its assigned treatment status.  There is still a dependence on the training set observations and their treatment assignment. This, however, is analogous to the analyst shaping the object of inquiry or creating a codebook after a pre-test.  With the AISV addressed, it is now possible to define key properties of the estimator, like bias or consistency.

The sample split also addresses the concerns about overfitting.  The analyst can explore in the training set as much as she likes, but, because findings are verified in a test set that is only accessed once, she is incentivized to find a robust underlying pattern. Patterns in the training set which are due to idiosyncratic noise are highly unlikely to also arise in the test set which helps assure the analyst that patterns which are confirmed by the separate test set will be replicable in further experiments.  By locking $g$ into place in the training set, the properties of the tests in the test set do not depend upon the number of different $g$'s considered in the training set.  In practice, we find splitting the sample ensures that we are able to consider several models to find the $g$ that best captures the data and aligns with our theoretical quantity of interest without worrying about accidentally p-hacking.

With the reason for sample splitting established, we first describe our final estimands for the text as outcome and text as treatment cases (Sections \ref{sub:outcome} and \ref{sub:treatment}).   We then describe the pragmatic steps we suggest to take to implement a train/test split (Section~\ref{sub:procedure}).  Then in Section~\ref{sub:tradeoffs}, we discuss the tradeoffs in using a split sample approach.  Having described our strategy, in Section~\ref{sub:priorwork}, we connect our approach to existing prior work before demonstrating how it works in two different applications (Section~\ref{sec:application}).

\subsection{Text as outcome}
\label{sub:outcome}

The text as outcome setting is straightforward.  The particular $g$ that the analyst chooses defines the categories of the outcome from which the estimand will be defined. Our goal is to obtain a consistent (and preferably unbiased) estimator for the ATE (or other causal quantities of interest) assuming a particular $g$.  Using Assumptions 1-3, a consistent estimator will be:

\begin{eqnarray}
\widehat{ATE} 
& = & \sum_{i \in \boldsymbol{I} } \frac{ I(T_{i} = 1) g_{\boldsymbol{J}}(\boldsymbol{Y}_{i} (1  )  )   }{\sum_{i \in \boldsymbol{I} } I(T_{i} = 1 ) }  -  \sum_{i \in \boldsymbol{I} } \frac{ I(T_{i} = 0) g_{\boldsymbol{J}}(\boldsymbol{Y}_{i} (0  )  )   }{\sum_{i \in \boldsymbol{I}} I(T_{i} = 0 ) } \nonumber
 \end{eqnarray}

When $g$ is fixed before documents $\boldsymbol{I}$ are examined, we can essentially treat the mapped outcome $g_{\boldsymbol{J}}(\boldsymbol{Y}_{\boldsymbol{I}})$ as an observed variable.\footnote{It is still important to verify that the mapped variable is capturing what you care about the underlying text.  Ultimately this is not any different than ensuring that a chosen outcome for an experiment captures the phenomenon of interest to the researcher.} Appendix~\ref{sub:proofs} gives an identification proof.

\subsection{Text as treatment}
\label{sub:treatment}

Text may also be the treatment in an experiment.  For example, we may ask individuals to read a candidate's biography and then evaluate how the candidate's favorability on a scale of 0 to 100. The treatment, $\boldsymbol{T}_i$, is the text description of the candidate assigned to the respondents.  The potential outcomes $Y_i(\boldsymbol{T}_i)$ describes respondent $i$'s rating of the candidate under the treatment assigned to respondent $i$.

While we could compare two completely separate candidate descriptions,  social scientists are almost always interested in how some underlying feature of a document affects responses---that is the researcher is interested in estimating how an \emph{aspect} or \emph{latent} value of the text influences the outcome.\footnote{This distinguishes our framework from A/B tests commonly found in industry settings which evaluate different blocks of text without attempting to understand why there are differences across the texts.}  
For example, the researcher might be interested in whether including military service in the description has an impact on the respondents' ratings of the candidate.  Military service is a latent variable -- there are many ways that the text could describe military service that all would count as the inclusion of military service and many ways that the text could omit military service that all would count as the absence of the latent variable. The researcher might assign 100 different candidate descriptions, some which mention the candidate's military service and some which do not.  In this case, the treatment of interest is $Z_i=g(T_i)$ which maps the treatment text to an indicator variable that indicates whether or not the text contains a description of the candidate's military service.  To estimate the impact of a binary treatment, we could use the estimator:

\begin{eqnarray}
\widehat{ATE} 
& = & \sum_{i \in \boldsymbol{I} } \frac{ I(Z_i=g_{\boldsymbol{J}}(\boldsymbol{T}_i)= 1) Y_{i} (1)  )   }{\sum_{i \in \boldsymbol{I} } I(Z_i=g_{\boldsymbol{J}}(\boldsymbol{T}_i)= 1 ) }  -  \sum_{i \in \boldsymbol{I} } \frac{ I(Z_i=g_{\boldsymbol{J}}(\boldsymbol{T}_i)= 0) Y_{i} (0  )  )   }{\sum_{i \in \boldsymbol{I} } I(Z_i=g_{\boldsymbol{J}}(\boldsymbol{T}_i)= 0 ) } \nonumber
 \end{eqnarray}

With text as treatment, we may be interested in more than just one latent treatment. The presence of multiple latent treatments requires different causal estimands and enables us to ask different questions about how features of the text affect responses.  For example, we can learn the marginal effect of military service and how military service interacts with other features of the candidate's background---such as occupation or family life.  Typically with multidimensional treatments we are interested in the effect of one treatment holding all others constant.  This complicates the use of topic models which suppose $\mathcal{Z}$ is a simplex (all topic proportions are non-negative and sum to one) because there is no straightforward way to change one topic holding others constant (see \citealt{fong2016discovery} and Appendix~\ref{app:binary}).  Instead we will work with $g$ that compress the text $\boldsymbol{T}$ to a vector of $K$ binary treatments $\boldsymbol{Z}_j \in \mathcal{Z}$ where $\mathcal{Z}$ represents all $2^K$ possible combinations of the treatments. We could also, of course, suppose that $g$ maps $\boldsymbol{T}$ to a set of continuous underlying treatments, but this requires additional functional form assumptions. 

The use of binary features leads naturally to the \emph{Average Marginal Component Effect} (AMCE), the causal estimand commonly used in conjoint experiments \citep{hainmueller2013causal}.  The AMCE estimates the marginal effect of one component $k$, averaging over the values of the other components:  
\begin{eqnarray*}
AMCE_k &=& \sum_{\boldsymbol{Z}_{-k}}E[Y(Z_k=1,\boldsymbol{Z}_{-k}) -  Y(Z_k=0, \boldsymbol{Z}_{-k})]m(\boldsymbol{Z}_{-k}) \nonumber 
\end{eqnarray*} 

The $AMCE_k$ describes the average effect of component $k$, summed over all other values of $k$, weighted by $m(Z_{-k})$, or an analyst determined distribution of $Z_{-k}$.  The AMCE can be thought of as an estimate of the effect of component $k$, averaging over the distribution of other components in the population---therefore providing a sense of how an intervention will matter averaging over other characteristics.

In order to discover the mapping from text to latent treatments we an additional assumption than in the text as outcome case.  This is because analysts are usually only able to randomize at the text level, but we are interested in identifying the effect of latent treatments we are unable to manipulate directly.  Consequently, we need to make an additional assumption beyond the three mentioned above in Section~\ref{sub:notation} (SUTVA, Ignorability and Positivity\footnote{To address the multidimensional treatments, the positivity assumption becomes the common support assumption which states that all combinations of treatments have non-zero probability $f(\boldsymbol{Z}_i) > 0 $ for all $\boldsymbol{Z}_i \in \text{Range}\; g(\cdot)$.}).The Sufficiency Assumption states that our $g$ captures all the information relevant to the response in $\boldsymbol{T}$ is contained in $\boldsymbol{Z}$

\citet{fong18explanatory} shows that for sufficiency to hold for any individual the response to two documents with the same latent feature representation might differ, but on average over individuals the responses are the same.  Mathematically, it is written as:\footnote{\citet{fong2016discovery} present a stronger and  more intuitive version. \citet{fong2016discovery} show that sufficiency holds if  $Y_{i}(\boldsymbol{T}_{i}) = Y_{i} (g(\boldsymbol{T}_{i}))$ for all documents and for all respondents.  In words, this assumption requires that the potential outcome response to the text be identical to the potential outcome response to all documents with the same latent feature representation.  This assumption is strong because it requires that there is no other information contained in the text that matters for the response beyond what is contained in the latent feature representation.  In our running example about military service, this would mean that the inclusion or exclusion of military service is the only aspect relevant to the effect of the document on the individual's rating.  Particularly for text, we could imagine that this assumption could easily be violated.  If both versions of the treatment contain ``The candidate served in the military'', but one also adds ``The candidate was dishonorably discharged'' we might expect that this additional text added in addition to $\boldsymbol{Z}$ may be relevant to the responses.}
\begin{ass}[Sufficiency]
For all $\boldsymbol{T}$ and $\boldsymbol{T}^{'}$ such that $g(\boldsymbol{T}) = g(\boldsymbol{T}^{'}) $ then $E[Y_{i}(g(\boldsymbol{T}))] = E[Y_{i}(g(\boldsymbol{T}^{'}))]$. 
\end{ass}  

\citet{fong18explanatory} shows that this assumption is equivalent to supposing that the components of the document that affect the response and are not included in the latent feature representation are orthogonal to the latent feature representation.  Technically, we can define $\epsilon_i(T) = Y_i(T) - Y_i(g(T))$ and then this more general assumption is equivalent to assuming that $E_i[\epsilon_i(T)] = 0$ for all $T$.  \citet{fong2016discovery} and \citet{fong18explanatory} provide an identification proof.  

\subsection{Procedure}
\label{sub:procedure}

In this section we discuss the general procedure for implementing the train/test split to estimate the above quantities of interest. This procedure follows the schematic in Figure~\ref{f:procedure}. Considerations specific to treatment or outcome are deferred to Appendix~\ref{sub:stm} and Appendix~\ref{sub:sibp}.

\subsubsection{Splitting the sample} The first major choice that the analyst faces is how to split the sample into two pieces: the training set and the test set.  A default recommendation is to split $50\%$ of the documents in training and $50\%$ in the test set.  But this depends on how the researcher evaluates the tradeoff between discovery of $g$ and testing.  Additional documents in the training set enables learning a more complicated $g$ or more precise coding rules.  Additional documents in the test set enable estimation of a more precise effect.  While the test set should be representative of the population that you want to make inference about, the training set can draw on additional non-representative documents as long as they are similar enough to the test set to aid in learning a useful $g$.  Finally, when taking the sample the analyst can stratify on characteristics of interest to ensure that the split has appropriate balance between the train and test set on those characteristics.  

Once the test set is decided, the single most important rule is that the test set is used once, solely for estimation.  If the analyst revises $g$ after looking at the test set data, she reintroduces the AISV and risks overfitting. Setting aside test data must be true for all features of the analysis: even preliminary steps like preprocessing must not include the test data set.  Third parties, such as survey firms and research agencies, can be helpful in credibly setting the data aside.  

\subsubsection{Discover $g$} We use the training set and text as data methods to find a $g$ that is interpretable, of theoretical interest, has high label  fidelity and is tractable.  In this paper we use the Structural Topic Model and the Supervised Indian Buffet Process but there are numerous other methods that are applicable.

\subsubsection{Validation in the training set} Validation is an important part of the text analysis process and researchers should apply the normal process of validation to establish label fidelity.  These validations are often application-specific and draw on close reading of the texts.\footnote{See \cite{grimmer2013text} for more detail on types of validation and the \texttt{stm} package \citep{roberts2017package} for tools designed to assist with validation.} These validations should be completed in the training set as part of the process of discovering and labeling $g$, before the test set is opened.

\subsubsection{Before opening the test set} While obtaining $g$ in the training set, we can refit $g$ as often as it is useful for our analysis.  But once applied to the test set we cannot alter $g$ further.  Therefore, we advise two final steps.  
\begin{itemize}
\item[1)] Take One More Look at $g$ \\
Be sure $g$ is capturing the aspect of the texts that you want to capture, assign labels and then validate to ensure that the conceptual gap between those labels and the representation $g$ produces is as small as possible.  While validation approaches may vary- this necessarily involves reading documents \citep{krippendorff2004content, quinn2010analyze, grimmer2013text}.  It is helpful to fix a set of human assigned labels, example documents and automated keyword labels in advance to avoid subtle influence from the test set.

\item[2)] Fix Your Evaluation Plan \\
While we focus on inference challenges with $g$, standard experimental challenges remain.  
Here we can draw from the established literature on best practices in experiments \citep{gerber2012field} potentially including a pre-analysis plan  \citep{humphreys2013fishing}. This can include multiple-testing and false-discovery rate corrections.
\end{itemize}

\subsubsection{Applying $g$ and estimating causal effects} Mechanically, applying $g$ in the test set is straightforward and is essentially the process of making a prediction for a new document.  After calculating the quantities $g_{\boldsymbol{J}}(\mathbf{Y_{\boldsymbol{I}}})$ we can use standard estimators appropriate to our estimand, such as the difference of means to estimate the average treatment effect.  The appendix describes how to apply $g$ to new documents in both the Supervised Indian Buffet Process and the Structural Topic Model, which we cover in our examples.

\subsubsection{Validation in the test set} It is also necessary to ensure that the model fits nearly as well on the test set as it did on the training set.  When both the training and test sets are random draws from the same population this will generally be true.  But overfitting or a small sample size can result in different model fit. The techniques used to validate the original model can be used in the test set as well as common measures of model fit such as log likelihood.  Unlike the validation in the training set, during the validation in the test set the analyst cannot return to make changes to the model. Nevertheless, validation in the test set helps the analyst understand the substantive meaning of what is being  estimated and provides guidance for future experiments. Formally, our estimand is defined in terms of the empirically discovered $g$ in the training set.  However, invariably the analyst making a broader argument indicated by the label.  Validation in the test set verifies that \textit{label fidelity} holds and that $g$ represents the concept in the test set of documents.

\subsection{Tradeoffs}
\label{sub:tradeoffs}

The train-test split addresses many of our concerns, but it is not without cost.  Efficiency loss is the biggest concern.  In a 50/50 train-test split, half the data is used in each phase of analysis, implying half the data is excluded from each step.  At the outset, it is difficult to assess how much data is necessary for either the training or the test set.  The challenge in setting the size of the test set is that the analyst does not yet know what the outcome (or treatment) will be when the decision is made on the size of the split. The problem in setting the size of the training set is that \textit{we don't know the power we need for discovery}. Alternatively, we could focus first on determining the power needed for estimation of an effect and then allocate the remaining data for discovery.  This can be effective, but it requires that we are able to anticipate characteristics of our discovered treatment or outcome.

\subsection{Prior work}
\label{sub:priorwork}
Our central contribution is a framework that characterizes how to make causal inferences with texts, identifies problems that arise when making those causal inferences, and the explanation of why sample splitting addresses these challenges.  There has been comparatively little work on causal inference with latent variables.  \citet{lanza2013causal} consider causal inference for latent class models but do not give a formal statement of identifying assumptions or acknowledge the set of concerns we identify as an analyst induced SUTVA violation.  \citet{volfovsky2015causal} present a variety of estimands and estimation strategies for causal effects where the dependent variable is ordinal.  They provide approaches based both on the observed data as well as latent continuous outcomes. \citet{volfovsky2015causal} express caution about the latent variable formulation due to identification concerns and the subsequent literature  \citep[e.g.,][]{lu2015sharp} has moved away from it.  Unfortunately, many of their strategies based directly on the observed outcomes are unavailable in the much higher dimensional setting of text analysis.  One notable exception is \citet{gill2015judicial} which evaluates the causal effect of gender on individual words in judicial decisions.

In contrast to the paucity of work on the problem we identify, our proposed solution: sample splitting, has a long history in machine learning.  There has been a growing exploration of the use of train-test splits in the social sciences as well as causal inference \citep{wager2017estimation, chernozhukov2017double, anderson2017split}.  It is the natural solution to this class of problems and we certainly do not claim to be the first to introduce the idea of train-test splits into the area.  Our approach is mostly closely related to prior work by \citet{fafchamps2017using} and \citet{anderson2017split} which both advocate a form of split samples to aid in discovery.

Our work is also part of a burgeoning literature on the use of machine learning algorithms to enhance causal inference \citep{van2011targeted, athey2015machine, bloniarz2016lasso, chernozhukov2017double, wager2017estimation}. Much of this work focuses on estimating causal parameters on observed data and addressing a common set of concerns such as estimation and inference in high-dimensional settings, regularization bias and overfitting.  Our work complements this literature by exploring the use of latent treatments and outcomes.  Many pieces in this area call for sample splits or cross-validation for estimation and inference, providing additional justification for our preferred approach \citep[see e.g.][]{chernozhukov2017double}.  In Appendix~\ref{sec:appadaptive} we discuss the connection between our work and related work in biostatistics. 

\section{Applications}
\label{sec:application}

We demonstrate how to make causal inferences using text in two applications: one where text is the outcome and one where text is the treatment.  Our procedure is inherently sequential.  We advocate both using a split sample design when analyzing an experiment and explicitly planning to run experiments again, in order to accumulate knowledge.  In each of the applications below we explicitly describe the discovery process when analyzing the data.  Although we use specific models, STM for text as outcome and sIBP for text as treatment, the process we describe here is general to any process for discovering $g$ from data. 

\subsection{Text as outcome: an experiment on immigration}
\label{sub:outcomeapp}

To first demonstrate how to use text as a response in a causal inference framework, we apply the structural topic model to open-ended responses from a survey experiment on immigration \citep{roberts2014structural}.  Specifically, we build on an experiment first introduced in \citet{cohen2004measuring} to assess how knowledge about an individual's criminal history affects respondent's preference for punishment and deportation.  These experimental results contribute to a large literature about Americans' preferences about immigrants and immigration policy (see \citealt{HaiHop14} for a review) and a literature on the punishments people view as appropriate for crimes \citep{CarDarRob02}.  Critically, in both conditions of our experiment an individual has broken the same law, entering the country illegally, but differs solely on past criminal history.  We therefore ask how someone's past criminal behavior affects the public's preference for future punishment and use the open-ended responses to gather a stated reason for that preference.   

To address this question we report the results from three iterations of a similar experiment.  With each experiment we report our procedure for choosing $g$ and the treatment effects in order to provide clarity and to demonstrate how the process described in Figure~\ref{f:procedure} works in practice. The first results are based on responses initially recorded in \cite{cohen2004measuring}. We use this initial set of responses to estimate an initial $g$ and to provide baseline categories for the considerations respondents raise when explaining why someone deserves punishment.  In a second experiment we build on \cite{cohen2004measuring}, but address issues in the wording of questions, expand the set of respondents who are asked to provide an open ended response, and update the results with contemporary data.  We then run a third experiment because we discovered our $g$ performed poorly in the test set of the second experiment.  We also used that opportunity to improve small features of the design of the experiment.   

We report the results of each experiment in order to be transparent about our research process, something we suggest that researchers do in order to avoid selective reporting based on an experiment's results.  The three sets of experimental results show that there has been surprising stability in the considerations Americans raise when explaining their punishment preferences, though there are some new categories that emerge.  There is also a consistent inclination to punish individuals who have previously committed a crime, even though they committed the same crime as someone without a criminal history.  

\subsubsection{Experiment 1}

As a starting point, we conduct an analysis of the results of an experiment reported in \citet{cohen2004measuring}.  The survey experiment was administered in the context of a larger study of public perceptions of the criminal justice system.  The survey was conducted in 2000 by telephone random-digit dial and includes 1,300 respondents.\footnote{More details about the survey are available in \citet{cohen2002measuring}.}  

In the experiment, respondents were given two scenarios of a criminal offense.  In both the treatment and control conditions, the same crime was committed: illegal entry to the United States.  In the treatment condition, respondents were told that the person had previously committed a violent crime and had been deported.  In the control condition, respondents were told that the person had never been imprisoned before.

In the treatment condition, respondents were told:

\begin{framed}
\begin{quote}
``A 28-year-old single man, a citizen of another country, was convicted of illegally entering the United States. Prior to this offense, he had served two previous prison sentences each more than a year. One of these previous sentences was for a violent crime and he had been deported back to his home country.''
\end{quote}
\end{framed}

In the control condition, respondents were told:
\begin{framed}
\begin{quote}
``A 28-year-old single man, a citizen of another country, was convicted of illegally entering the United States. Prior to this offense, he had never been imprisoned before.''
\end{quote}
\end{framed}

Respondents were then asked a close-ended question about whether or not the person should go to jail.  If they responded that the person should not go to jail, they were asked to respond to an open-ended question, ``Why?"   The key inferential goal of the initial study was determining if a respondent believed a person should be deported, jailed, or given some other punishment.

\subsubsection{Experiment 2}

After analyzing the results of Experiment 1, we ran a second experiment using the same treatment and control conditions, but with slight design differences to build upon and improve the original experimental protocol.  First, all respondents were asked the open-ended question, not just those who advocated for not sending the individual to jail.  Second, we redesigned the survey to avoid order effects.  Third, we asked a more specific open-ended question.  We still asked `Should this offender be sent to prison?' (responses: yes, no, don't know) but followed by asking ``Why or why not? Please describe in \textbf{at least two sentences} what actions if any the U.S. government should take with respect to this person and why?''\footnote{Per our IRB we added the statement ``(Please \textbf{do not} include any identifying information such as your name or other information about you in this open-ended response.)''}  Experiment 2 was run on Mechanical Turk on July 16, 2017 with 1000 respondents.

\subsubsection{Experiment 3}

We expected Experiment 2 to be our last experiment, but we encountered a design problem.  After we estimated $g$ in the training set using STM and fit it to the test data, we realized that some of our topic labels were inaccurate.  In particular, we had attempted to label topics using three pre-determined categories: prison, deport, and allow to stay.  But the data in the second experiment suggested some new categories.  We could not simply relabel the topics in the test set, because this would eliminate the value of the train/test split.  Instead we verified the results of experiment 2 with an additional experiment.\footnote{We also took the opportunity to make a few design changes.  We had previously included an attention check which appeared after the treatment question.  We moved the attention check to before the treatment.  We also had not previously used the MTurk qualification enforcing the location to be in the U.S. although we did in Experiment 3.  Finally, we blocked workers who had taken the survey in Experiment 2 using the \texttt{MTurkR} package \citep{leeper2017}.}  Experiment 3 was run on Mechanical Turk on September 10, 2017 with 1000 respondents.  To avoid labeling mistakes, two members of our team labeled the topics independently using the training data and then compared labels with one another to create a final set of congruent labels before applying the $g$ to the test set.

\subsubsection{Results}

In each experiment, we used equal proportions of the sample in the train and test sets. In each experiment we fit several models in the training set before choosing a single model that we then applied to the test set.  

We include the results from all three experiments below, though because of space constraints we put a description of topics and representative documents of Experiments 1 and 2 in the Appendix.  For Experiment 3, Table \ref{topiclabels3} shows the words with the highest probability in each of 11 topics and the documents most representative of each topic, respectively.  Topics range from advocating for rehabilitation or assistance for remaining in the country to suggesting that the person should receive maximal punishment.

\begin{table}[ht!!]
\centering
\begin{scriptsize}
\begin{tabular}{r p{.4\textwidth} p{.4\textwidth}}
  \hline
 & Label & Highest Probability Words \\ 
  \hline
Topic 1 & Limited punishment with help to stay in country, complaints about immigration system & legal, way, immigr, danger, peopl, allow, come, countri, can, enter \\ 
  Topic 2 & Deport & deport, think, prison, crime, alreadi, imprison, illeg, sinc, serv, time \\ 
  Topic 3 & Deport because of money & just, send, back, countri, jail, come, prison, let, harm, money \\ 
  Topic 4 & Depends on the circumstances & first, countri, time, came, jail, man, think, reason, govern, put \\ 
  Topic 5 & More information needed & state, unit, prison, crime, immigr, illeg, take, crimin, simpli, put \\ 
  Topic 6 & Crime, small amount of jail time, then deportation & enter, countri, illeg, person, jail, deport, time, proper, imprison, determin \\ 
  Topic 7 & Punish to full extent of the law & crime, violent, person, law, convict, commit, deport, illeg, punish, offend \\ 
  Topic 8 & Allow to stay, no prison, rehabilitate, probably another explanation & dont, crimin, think, tri, hes, offens, better, case, know, make \\ 
  Topic 9 & No prison, deportation & deport, prison, will, person, countri, man, illeg, serv, time, sentenc \\ 
  Topic 10 & Should be sent back & sent, back, countri, prison, home, think, pay, origin, illeg, time \\ 
  Topic 11 & Repeat offender, danger to society & believ, countri, violat, offend, person, law, deport, prison, citizen, individu \\ 
   \hline
\end{tabular}
\end{scriptsize}
\caption{Experiment 3: Topics and highest probability words}
\label{topiclabels3}
\end{table}

\begin{figure}
\includegraphics[scale=.7]{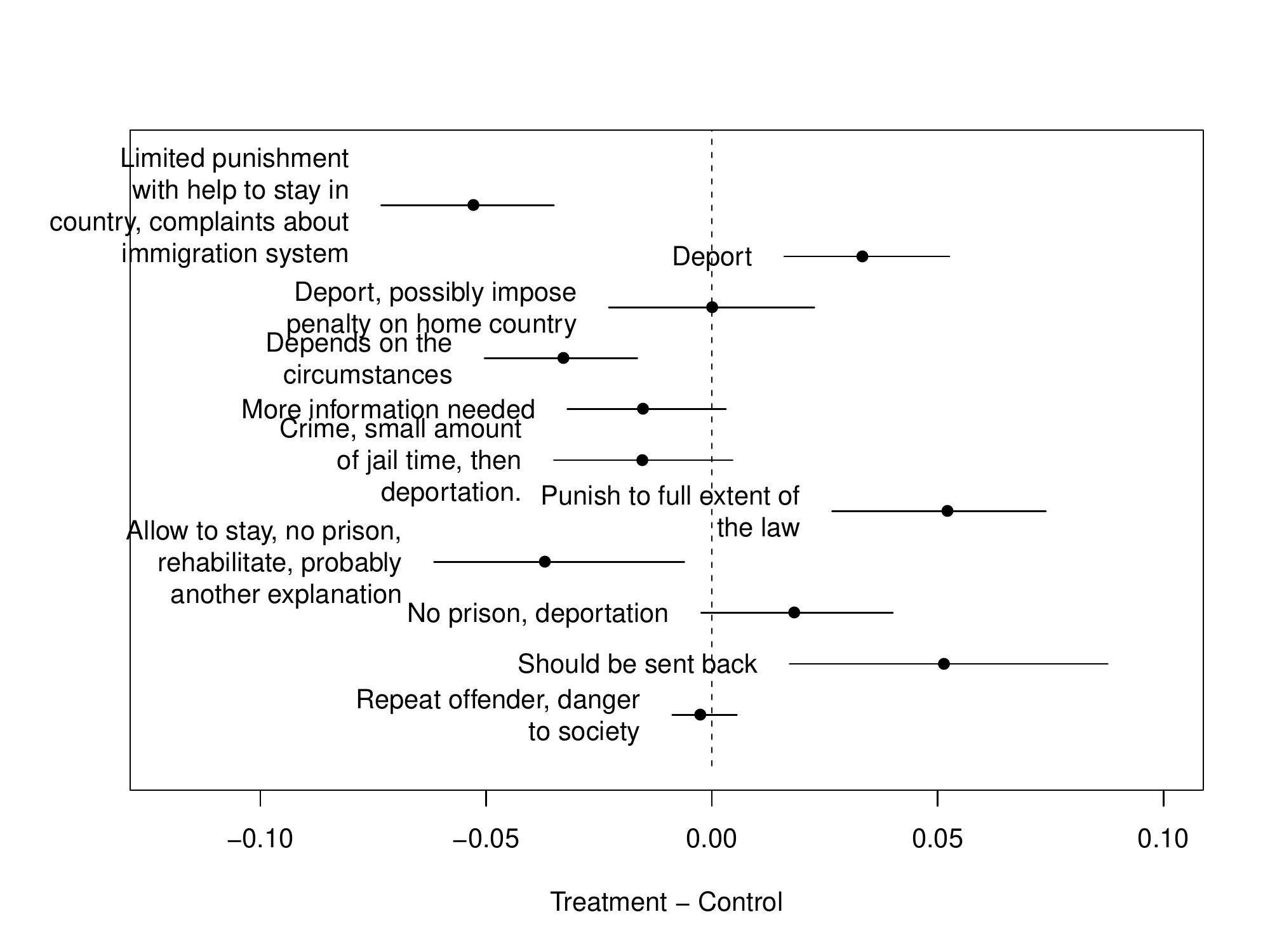}
\caption{Test Set results for Immigration Experiment 3.  Point estimates and $95\%$ confidence intervals. \label{immigrationresults}}
\end{figure}

After discovering, labeling, and finalizing $g$ in the training set, we estimated the effect of treatment on the topics in the test set.  In Figure \ref{immigrationresults} we show large impacts of treatment on topics.  Treatment (indicating that the person had a previous criminal history) increased the amount of writing about maximal punishment, deportation, and sending the person back to their country of origin.  The control group was more likely to advocate that the person should be able to stay in the country or that the punishment should depend on the circumstances of the crime.

\begin{figure}
\includegraphics[scale=.4]{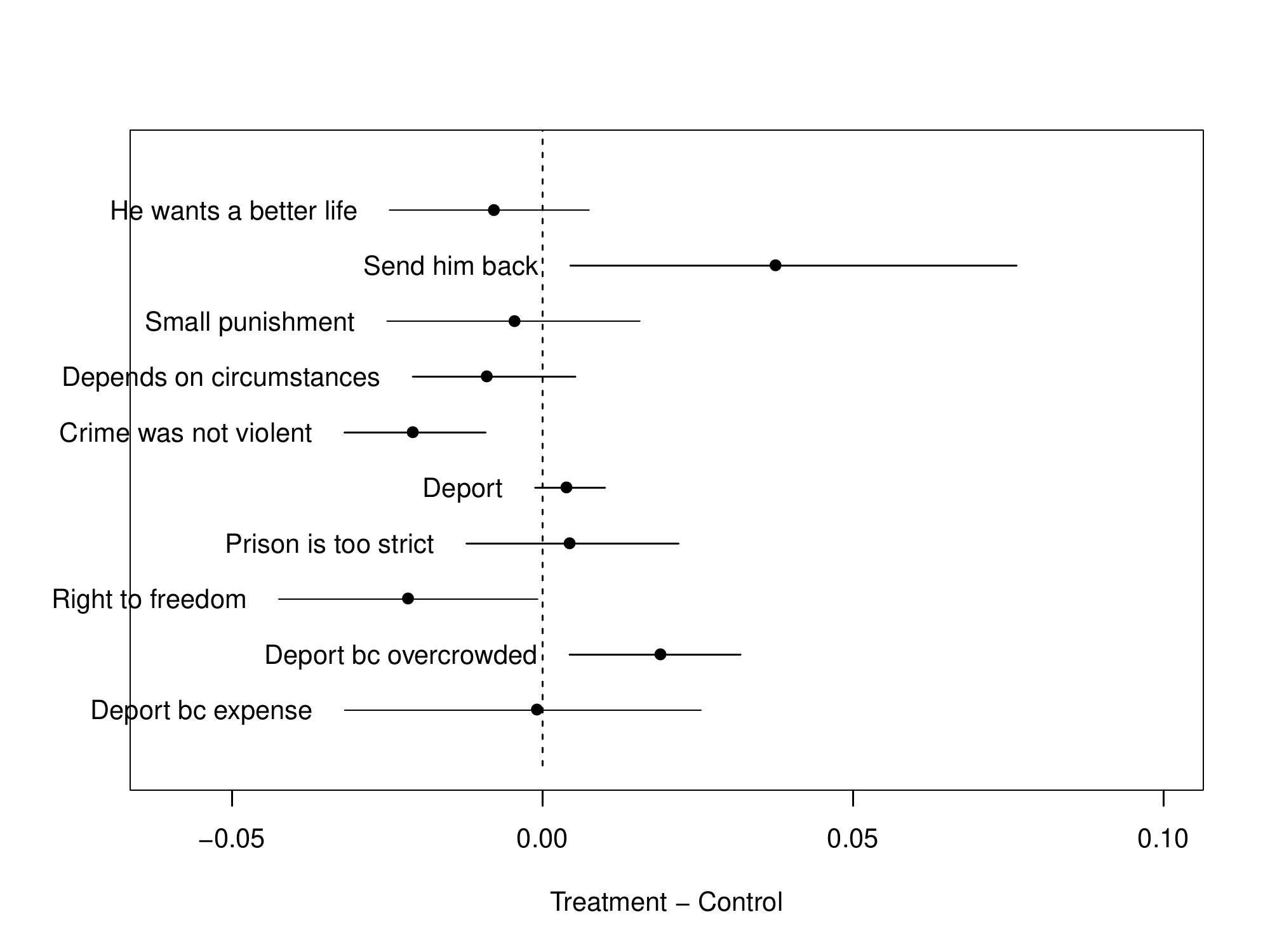}\includegraphics[scale=.35]{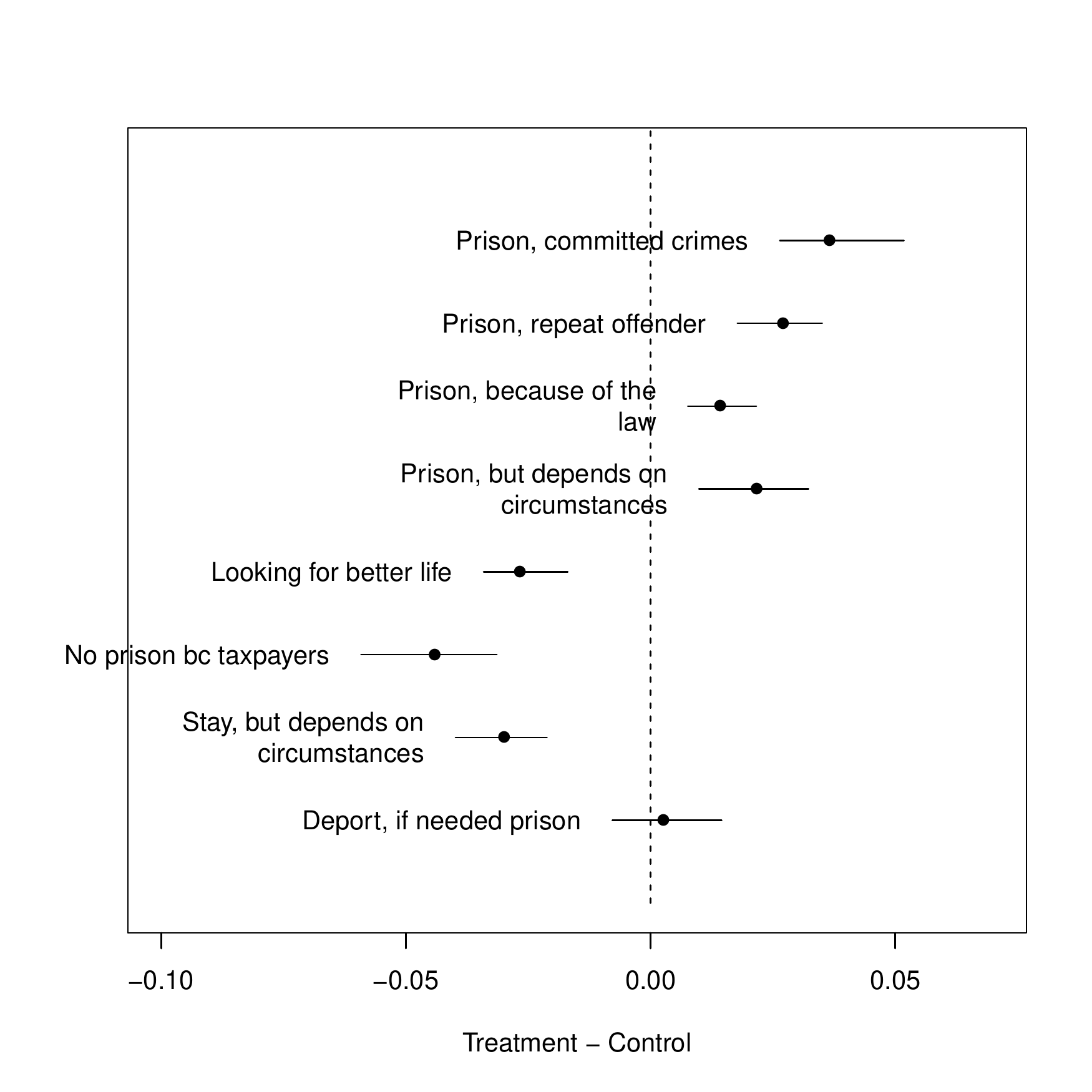} 
\caption{Test Set results for Experiment 1 (left) and Experiment 2 (right).  Point estimates and $95\%$ confidence intervals. \label{results1and2}}
\end{figure}

We found qualitatively similar results in Experiments 1 and 2 (Figure \ref{results1and2}), even though $g$ is different in both cases and the set of people who were asked to provide a reason is different.  In each case, the description of a criminal history significantly increases the likelihood that the respondent advocates for more severe punishment or deportation.

\paragraph{Next Steps} In Figure~\ref{f:procedure}, we recommend concluding experiments with suggestions for further experimentation and we do so here.  Future iterations of the experiment could explore two features of the treatment.  First, we have only provided information about one type of crime.  It would be revealing to know how individuals respond to crimes of differing severity.  Second, we could use our existing design to estimate heterogeneous treatment effects, which would be particularly interesting in light of contemporary debates about how to handle undocumented immigration in the United States.

\subsection{Text as treatment: Consumer Financial Protection Bureau}
\label{sub:treatmentapp}
We turn next to examine how our framework applied to text-based treatments.  We examine the features of a complaint that causes the Consumer Financial Protection Bureau (CFPB) to reach a timely resolution of the issue.  The CFPB is a product of Dodd-Frank legislation and is (in part) charged with offering protections to consumers.  The CFPB solicits complaints from consumers across a variety of financial products and then addresses those complaints. It also has the power to secure payments for consumers from companies, impose fines on firms found to have acted illegally, or both. 

The CFPB is particularly compelling for our analysis because it provides a massive database on the text of the complaint from the consumer and how the company responded.  If the person filing the complaint consents, the CFPB posts the text of the complaint in their database, along with a variety of other data about the nature of the complaint.  For example, one person filed a complaint stating that 
\begin{framed}
\begin{quote}
the service representative was harsh and not listening to my questions. Attempting to collect on a debt I thought was in a grace period ...They were aggressive and unwilling to hear it
\end{quote}
\end{framed}
\noindent and asked for remedy.  The CFPB also records whether a business offers a timely response once the CFPB raises the complaint to the business.  In total, we use a collection of 113,424 total complaints downloaded from the CFPB's public website.

The texts are not randomly assigned to the CFPB, but we view the use of CFPB data as still useful for demonstrating our framework.  Much of the information available to bureaucrats at the CFPB will be available in the complaint, because of the way complaints are recorded in the CFPB data.  To be clear, for the effect of the text to be identified, we would need to assume that the texts provide all the information for the outcome and that any remaining information is orthogonal to the latent features of the text.  We view the example of the CFPB as useful, because it provides us a clear way to think through how this assumption could be violated. If there are other non-textual factors that correlate with the text content, then our estimated treatment effects will be biased.  For example, if working with the CFPB directly to resolve the complaint were important and individuals who submitted certain kinds of complaints were less well equipped to assist the CFPB, then we would be concerned about whether selection on observables holds.  Or, there could be demographic factors that confound the analysis.  For example, minorities may receive a slower response from CFPB bureaucrats or a more adversarial response from financial institutions \citep{Butler14,Costa17} and minorities may be more likely to write about particular topics.  While this is certainly plausible, many of the effects that we estimate of the text are large, so they would be difficult to explain solely through this confounding.  

Our goal is to discover the treatments and estimate their effect on the probability of a response.  We discover $g$ using the supervised Indian Buffet Process developed for this setting in \cite{fong2016discovery} and implemented in the \texttt{texteffect} package in \texttt{R} \citep{texteffect}.  The model learns a set of latent binary features which are predictive of both the text and the outcome.  To do this, we first randomly divide the data, placing 10\% in the training set and 90\% of the data in the test set.  We place more data in the test set because our large sample ($\approx$ 11K) provides ample opportunity to discover the latent-treatments in the training set and to provide greater power when estimating effects in the test set.  In the training set we apply the sIBP to the text of the complaints and whether there was a timely response.  We use an extensive search to determine the number of features to include and the particular model run to use.  The sIBP is a nonparametric Bayesian method; based on a user-set hyperparameter, it estimates the number of features to include in the model, though the number estimated from a nonparametric method rarely corresponds to the optimal number for a particular application.  To select a final model we then evaluate the candidate model fits utilizing a model fit statistic introduced in \cite{fong2016discovery} that provides a quantitative measure of model fit.  The train/test split ensures that we can refit the model several times choosing the estimate that provides the features that provide the best substantive insights.

Once we have fit the model in the training set, we use it to the infer the treatments in the test set.  Table \ref{t:latent} provides the inferred latent treatments from the CFPB complaint data.  The \emph{Automatic Keywords} are the words with the largest values in the estimated latent factors for each treatment, and the manual keyword is a phrase that we assign to each category after assessing the categories.  Using these features we can then infer their presence or absence in the treated documents and then estimate their effect.  To do this we use the regression procedure from \cite{fong2016discovery} and then use a bootstrap to capture uncertainty from estimation.

\begin{table}
\caption{Consumer Financial Protection Bureau Latent Treatments}\label{t:latent}
\centering
\begin{scriptsize}
\begin{tabular}{l|l|l}
\hline\hline
No. & Automatic Keywords & Manual Keyword \\
\hline
1  & payment, payments, amount, interest, balance, paid, month  & loan \\
2 & card, called, call, branch, money, deposit, credit\_card, told & bank \\
3 & debt, debt\_collection, account, number, validation, dispute, collection & debt collection \\
4 & xxxx, account, time xxxx\_xxxx, request, copy, received, letter & detailed complaint\\
5 & payment, payments, pay, told, amount, month, called & disputed payment \\
6 & loan, mortgage, modification, house foreclosure, payments & mortgage\\
7 & debt, debt\_collection, collection, credit\_reporting, proof, credit\_report & threat \\
8 & fcra, credit\_report, credit\_reporting, reporting, debt, violation, law & credit report \\
\hline\hline
\end{tabular}
\end{scriptsize}
\end{table}

\begin{figure}
\caption{The Effect of Complaint Features on a Prompt Response} \label{f:latent}
\begin{center}
\scalebox{0.8}{\includegraphics{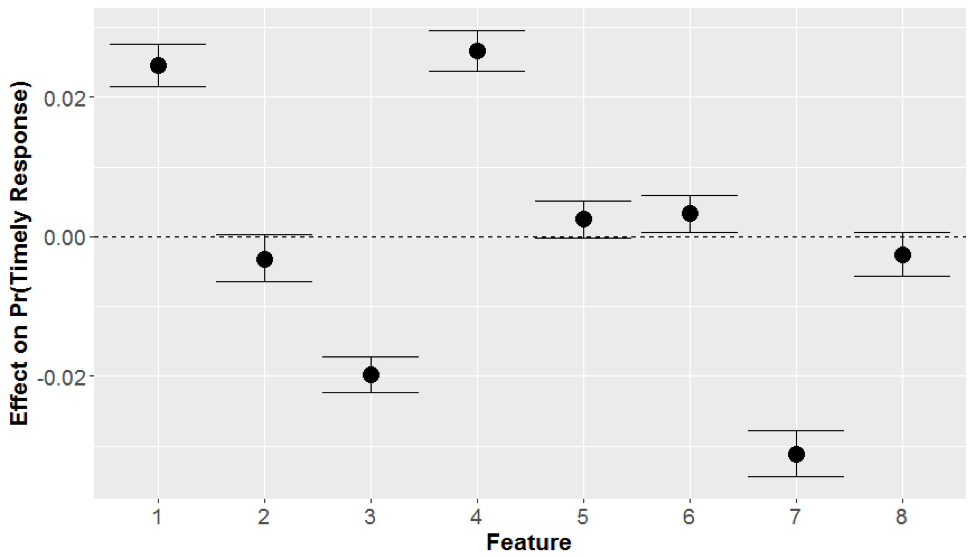}}
\end{center}
\end{figure}

Figure \ref{f:latent} shows the effects of each latent feature on the probability of a timely response.  The black dots are point estimates and the lines are 95-percent confidence intervals.  Figure \ref{f:latent} reveals that when consumers offer more detailed feedback (Treatment 4) and when complaints are made about payments to repay a loan (Treatment 1), the probability of a prompt response increases.  In contrast, the CFPB is much less successful at obtaining prompt responses from debt collectors---either when those collectors are explicitly attempting to collect a debt (Treatment 3) or when the debt collectors are threatening credit reports (Treatment 7).  The inability to obtain a prompt response from debt collectors is perhaps not surprising---debt collection companies exist to successfully recover funds and are likely less concerned with their perceived reputation with debtors.  It also demonstrates that it can be harder to remedy consumer complaints in some areas than others, even if the CFPB is generally able to assist complaints.

\paragraph{Next Steps}  If we were to run a further iteration of the CFBP analysis, we would proceed on two fronts.  First, there is a constant stream of data arriving at the CFPB.  We could use our existing $g$ to reestimate the training effects to see if there are temporal trends.  We could also estimate a new $g$ to assess if new categories emerge over time.  Second, we could design experiments to address concerns about demographic differences.  For example, we could partner with individuals who are planning to write complaints to see how their language, independent of their personal characteristics, affects the response.    

\section{Conclusion}
\label{sec:conclusion}

Text is inherently high-dimensional.  This complexity makes it difficult to work with text as an intervention or an outcome without some simplifying low-dimensional representation.  There are a whole host of methods in the text as data toolkit for learning new, insightful representations of text data.  Unfortunately, while these low-dimensional representations make text comprehensible at scale, they also make causal inference with text difficult to do well, even within an experimental context.  When we discover the mapping between the data and the quantities of interest, the process of discovery undermines the researcher's ability to make credible causal inference.

In this paper we have introduced a conceptual framework for causal inference with text, identified new problems that emerge when using text data for causal inference, and then described a procedure to resolve those problems.  In this conceptual framework, we have clarified the central role of $g$, the codebook function, in making the link between the high-dimensional text and our low-dimensional representation of the treatment or outcome.  In doing so we clarify two threats to causal inference: the Analyst-induced SUTVA violation---an identification issue--- and overfitting---an estimation issue.  We demonstrate that both the identification and estimation concerns can be addressed with a simple split of the dataset into a training set used for discovery of $g$ and a test set used for estimation of the causal effect.  More broadly, we advocate for research designs that allow for sequential experiments that explicitly set aside research degrees of freedom for discovery of interesting measures, while rigorously testing relationships within experiments once these measures are defined explicitly.

Our conceptual framework and procedure unifies the text as data literature with the traditional approaches to causal inference.  We have considered the text as treatment and text as outcome, and in the future we hope to address the setting of text as treatment and outcome.  In related work, \citet{roberts2017matching} consider the text-based confounding setting. There is much more work to be done to explore other causal designs and improvements on the work we have presented here including optimally setting training/test splits and increasing the efficiency of discovery methods so that they can work on even smaller data sets.  

While our argument has principally been about the analysis of text data, our work has implications for any latent representation of a treatment or outcome used when making a causal inference. This could include latent measures common in political science such as measures of democracy (e.g. Polity), voting behavior (e.g. ideal points) and forms of manual content analysis.  Any time a process of discovery is necessary, we should be concerned if the discovery is completed on the same units where the effect is estimated.  In certain circumstances this process will be unavoidable.  Polity scores were developed by looking at the full population of world democracies so there is no test set we can access, but we argue that the train/test split should be considered in the context of the development of future measures that require a low-dimensional representation of high-dimensional data.

What do our findings mean for existing applied work (text and otherwise)? The AISV and overfitting raise considerable risks to replicability but it does not mean any work not employing a train-test split is invalid.    However, as estimands based on latent constructs become more common in the social sciences, we hope to see an increased use of the train-test split and the development of new methodologies to enhance the process of discovery.

\bibliographystyle{apsr}
\bibliography{ais}
\clearpage

\appendix

\section{Online Appendix}
\label{sec:techappendix}

This appendix expands on the main paper, filling in a number of specific details.  
\begin{itemize}
\item Section~\ref{app:techsection} contains proofs and additional technical clarifications alluded to in the main text.  
\item Section~\ref{sec:appadaptive} draws out additional connections to the machine learning literature.  
\item Section~\ref{app:procedure} outlines the procedure and clarifies variance estimation.  
\item Section~\ref{sub:stm} and Section~\ref{sub:sibp} provide details for STM and SIBP respectively.  
\item Section~\ref{app:stability} assess stability of the STM model across train and test splits.  
\item Section~\ref{app:experiment} collects additional materials from the experiments reported in Section~\ref{sub:outcomeapp} of the main paper.
\end{itemize}

\subsection{Proofs and Technical Details}
\label{app:techsection}
\subsubsection{Estimation with a true $g$}
\label{fn:gcheck}
It might seem natural to inquire about the properties of the estimator we use to obtain $g$.  In this setting, we can use the procedure to obtain $g$ as an estimator $G$. If we suppose that there is some true function $\check{g}$ we might ask how well our estimator $G$ performs---in large samples does the $g$ converge to $\check{g}$ and in small samples how discrepant is $g$ compared to $\check{g}$? 

While it is certainly useful to conceive of the estimator $G$, it is misguided to suppose that there is a true $\check{g}$ for some data set that our procedure is attempting to reveal.  To see why it is not useful to suppose there is a true function $\check{g}$ consider a hypothetical experiment where we examine how people respond to a knock on the door and encouragement to vote.  We might be immediately interested in whether respondents are more likely to express a positive tone about political participation.  To investigate this, we might construct a $g$ that measures the tone of open-ended responses.  But, we might also be interested in the topics that are discussed after receiving a mobilization, or whether individuals mention privacy concerns.  There is also large variation in the ways we might examine how the particular contents of the mobilization message might affect respondents.  We might be interested in whether messages that have a positive tone are more likely to increase turnout, whether highlighting the threats from a different political party causes an increase in turnout, or whether threatening the revelation of voter history to neighbors is the most effective method of increasing turnout.  This hypothetical example makes clear that there is no ``true" application-independent function for obtaining either the dependent variable or treatment when making causal inferences from texts.  Further, the fact that we need to discover $g$ at all implies that as the researcher we might be unsure about what properties we want $g$ to have---making it particularly difficult to evaluate the estimator \emph{a priori}.  

\subsubsection{Proof: Identifying ATE with text as dependent variable}
\label{sub:proofs}
This appendix section proves that after using the codebook function $g$ on text as a dependent variable the ATE is still preserved.  We then weaken conditions needed on $g$ to identify the ATE. 

We make Assumption 1-3 and we suppose that we have a codebook function $g$.  Without loss of generality we will suppose that the codebook function maps text into a set of $K$ categories with the constraint that the sum across all categories is equal to 1.  One example of this is using an STM to estimate the dependent variables from a set of texts.  Suppose further that we are interested in the effect of a dichotomous intervention on the prevalence of the $k^{\text{th}}$ category.  Our formal estimand of interest, then, is:

\begin{eqnarray}
\text{ATE}_{k} & = & \text{E}[z_{i,1, k} - z_{i,0,k} ]. \nonumber 
\end{eqnarray}

Where $z_{i,1,k}$ corresponds to the prevalence of the $k^{\text{th}}$ category for observation $i$ after receiving $T_{i} = 1$.  

We can see that the treatment effect is still identified by noting that after our randomization we have 

\begin{eqnarray}
&& \text{E}[g(\boldsymbol{Y}_{i}(T_{i} = 1)) | T_{i} = 1 ] - \text{E}[g(\boldsymbol{Y}_{i}(T_{i} = 0)) | T_{i} = 0 ] \nonumber \\
& = & \text{E}[z_{i,1, k} | T_{i} = 1 ] - \text{E}[z_{i,0, k}| T_{i} = 0 ] \nonumber \\
& = & \text{E}[z_{i,1, k} - z_{i,0, k}] = \text{ATE}_{k} \nonumber 
\end{eqnarray}

Where we apply the definition of $g$ and the randomization of the treatments.  Note that for this proof to work, it is essential that $g$ is fixed, otherwise the expectation is undefined. 

We can make a slightly weaker requirement of $g$ and still preserve identification of the causal effect.  Specifically, the only requirement is that any potential other $g$, $\tilde{g}$ agrees with $g$ for category $k$ for all text documents, or that $\tilde{g}(\boldsymbol{Y})_{k} = g(\boldsymbol{Y})_{k}$ for all $\boldsymbol{Y} \in \boldsymbol{\mathcal{Y}}$.  This implies the other categories could be arbitrarily different, but logically it requires that the total proportion of documents placed in the other $K-1$ categories is equal for both functions.  The proof follows immediately from the (obvious) proof above.  

\subsubsection{Technical Definition of AISV}
\label{app:technical}
In this section we offer a formal definition of the Analyst-Induced SUTVA Violation.  To formally define the AISV we rewrite $g$ as explicitly dependent on training data: both treatments $\boldsymbol{T}_{\boldsymbol{J}}$ and responses $\boldsymbol{Y}_{\boldsymbol{J}}$.  Specifically, we will write the value of $g_{\boldsymbol{J}}$ for observation $i$ that received treatment $\boldsymbol{T}_{i}$ as $g(\boldsymbol{Y}(\boldsymbol{T}_{i}), \boldsymbol{Y}_{\boldsymbol{J}}(\boldsymbol{T}_{\boldsymbol{J}}))$ where $\boldsymbol{Y}_{\boldsymbol{J}}(\boldsymbol{T}_{\boldsymbol{J}})$ describes all respondents' text-based responses and the vector of treatments for everyone in the set $\boldsymbol{J}$.  Suppose now that we re-randomize treatment $\boldsymbol{T}^{'}_{\boldsymbol{J}}$, such that $\boldsymbol{T}_{i} = \boldsymbol{T}_{i}^{'}$ and that $\boldsymbol{T}_{j} \neq \boldsymbol{T}_{j}^{'}$ for at least one $j \in \boldsymbol{J}\setminus i$.  Further, suppose we obtain new responses $\boldsymbol{Y}_{\boldsymbol{J}}(\boldsymbol{T}^{'}_{\boldsymbol{J}})$. 

AISV problems emerge if $g_{\boldsymbol{J}}(\boldsymbol{Y}(\boldsymbol{T}_{i}))  = g(\boldsymbol{Y}(\boldsymbol{T}_{i}), \boldsymbol{Y}_{\boldsymbol{J}}(\boldsymbol{T}_{\boldsymbol{J}})) \neq g(\boldsymbol{Y}(\boldsymbol{T}_{i}^{'}), \boldsymbol{Y}_{\boldsymbol{J}}(\boldsymbol{T}^{'}_{\boldsymbol{J}})) = g_{\boldsymbol{J}}(\boldsymbol{Y}(\boldsymbol{T}_{i}^{'}))$, even though $\boldsymbol{Y}(\boldsymbol{T}_{i}) = \boldsymbol{Y}(\boldsymbol{T}_{i}^{'})$---in plain language, the lower dimensional representation of document $i$ is different between the two randomizations even though the texts themselves are the same.  This is particularly problematic if we wanted to characterize the bias in estimators, or their properties in large samples.  This is because expectations are taken over different treatment allocations.  And different treatment allocations, under many different procedures for obtaining a codebook function $g$, imply that there are new categories of the dependent variable or new treatments in the text. 

\subsubsection{Assuming the AISV Away}
\label{app:away}
Formally, to assume away the AISV we would assume that $g_{\boldsymbol{J}}(\boldsymbol{Y}(\boldsymbol{T}_{i}), \boldsymbol{Y}_{\boldsymbol{J}}(\boldsymbol{T}_{\boldsymbol{J}})) = g_{\boldsymbol{J}}(\boldsymbol{Y}(\boldsymbol{T}_{i}^{'}), \boldsymbol{Y}_{\boldsymbol{J}}(\boldsymbol{T}^{'}_{\boldsymbol{J}}))$ for all $\boldsymbol{T}_{\boldsymbol{J}}, \boldsymbol{T}^{'}_{\boldsymbol{J}}$ and all $\boldsymbol{J}$.  However, the conditions for this stability can be surprisingly difficult to obtain \citep{chernozhukov2017double}. Assuming AISV away also does not solve the problem of overfitting.

\subsection{Further Connections to Literature}
\label{sec:appadaptive}
In this section we provide a further connection to the machine learning literature.  To make the connection, we compare our sequential approach to other methods for ensuring that we avoid overfitting.  One natural approach would be to adopt a cross-fitting or cross-validation approach which has been extremely successful in other contexts \cite{anderson2017split,chernozhukov2017double}.  In $k$-fold cross validation the data is partitioned into $k$ equally sized partitions.  The model is trained on all but one of these partitions (called the held-out set) and then model is estimated on the held-out set.  Then the procedure is repeated so each of the $k$ partitions is treated as the held-out set at least once.  This forms an estimate for every observation $i$ where the prediction comes from a model which was not trained on observation $i$.   This is a powerful approach but relies on the idea that the predictions will be comparable across observations which is true, for example, in settings where the estimand is well-defined in advance of the split.  In our setting, though, we have two problems that preclude the use of cross validation.  First, when a human is in the loop there is not way to separate the model fitting procedures because the human will remember the insights from the previous train-test split.  Second, because the estimand is not defined in advance of the split, every fold of the cross-validation could result in our procedure could result in us measuring slightly different concepts.  The result is that we would have no coherent way to align the $g$ across the cross validation folds.  Taken together, this suggests that a cross-validation or cross-fitting strategy could only be pursued under strong assumptions about the existence of a true $g$ or with severe limitations on the discovery process.  

\subsection{Explanation of Procedure}
\label{app:procedure}
The following steps are a road map for our procedure.  
\begin{itemize}
\item[1)] \textbf{Collect }a set of documents and split them into a training set and a test set.  Do not look at the test set.
\item[2)] Using your training set only, \textbf{choose} $g$ that compresses the high-dimensional text to a low-dimensional variable that will serve as either your treatment or outcome. Assign labels to low-dimensional categories.
\item[3)] \textbf{Validate} that the chosen $g$ accurately maps to a concept of theoretical significance for your argument.
\item[4)] \textbf{Estimate} the causal effect using the test set with the $g$ discovered in test set.  You can only use the test set once.
\item[5)] \textbf{Validate} that the $g$ worked as expected in the test set.
\item[6)] Ideally, \textbf{replicate} your findings in a new sample, repeating steps 1-5.  If you are unable to replicate, clarify what you would alter in the next experiment.  
\end{itemize}

\subsection{Uncertainty Estimation with $g$}
Once we have applied $g$ to our test data we can calculate confidence intervals using usual variance estimators that capture uncertainty about our estimate given a limited sample size conditional on $g$.  Examples in prior work tends to explicitly take the view of $g(\mathbf{Y})$ as a latent variable about which there is some additional measurement uncertainty and advocated approaches to incorporate this additional uncertainty into our confidence intervals \citep{roberts2014structural, fong2016discovery}.  For example, \cite{roberts2014structural} advocates a simulation approach to integrate over the variational approximation to the posterior distribution which conditions on the learned topic-word distribution, but accounts for the fact that the document-topic proportion $\mathbf{\theta}$ cannot be known with certainty for a particular document because it has a finite length.  \cite{fong2016discovery} use a bootstrap approach which captures measurement uncertainty both in the topic-word parameters and the document-topic representation.  While this approach is intuitively appealing, it complicates the definition of $g$ as a function because we run the risk of the same text mapping to two different values of the latent variable (failing the vertical line test).  In the interest of simplicity we do not include this form of measurement error in this article and leave to future work the incorporation of this uncertainty into the causal framework.

\subsection{Structural Topic Model}
\label{sub:stm}
The Structural Topic Model is a mixed membership model of texts related to Latent Dirichlet Allocation \citep{blei2012probabilistic} which was developed in \citet{roberts2014structural, roberts2016model} and implemented in the \texttt{stm} package in \texttt{R} \citep{ roberts2017package}.   It allows for the analyst to incorporate observed document metadata which is able to affect either topical prevalence (the amount which a topic is discussed) and topical content (the way in which a topic is discussed).  In this paper we consider the case in which a set of observed metadata which includes the treatment and pre-treatment covariates are allowed to affect topic prevalence and there are no topical content covariates.  Denoting the pretreatment covariates for document $i$ as $\boldsymbol{X}_i$ and the scalar treatment as $T_i$, the generative process can be given as:
\begin{align*}
\boldsymbol{\eta}_i &\sim \text{Normal}(\boldsymbol{X}_i\boldsymbol{\gamma}_X + T_i\gamma_{\tau}, \boldsymbol{\Sigma}) \\
 \theta_{i,k} &= \frac{\text{exp}(\eta_{i,k})}{\sum_{k=1}^K \text{exp}(\eta_{d,k})} \\
 z_{i,n} &\sim \text{Categorical}(\boldsymbol{\theta}_i) \\
w_{i,n} &\sim \text{Categorical}(\boldsymbol{\beta}_{z_{i,n}})
\end{align*}
Where $\boldsymbol{\theta}_d$ is a $K$-dimensional vector on the simplex indicating the proportion of the document allocated to each topic formed by applying the softmax function to $\boldsymbol{\eta}_d$ a vector in $\mathcal{R}^{K-1}$ where the $K$-th element is fixed to zero. $z_{i,n}$ is a token level latent variable containing the assignment for token $n$ of document $i$.  $\boldsymbol{\beta}$ is a $K$ by $V$ dimensional matrix where each row contains the conditional probability of seeing word $v$ given that is about topic $k$.  The model differs from Latent Dirichlet Allocation in its use of a logistic normal prior distribution for the document-topic proportions and through the ability to have that prior centered at a document-specific location determined by the document metadata. 

The model is estimated using partially-collapsed, non-conjugate, variational inference.  $\boldsymbol{\gamma}$ and $\boldsymbol{\Sigma}$ are given regularizing priors of the user's choice and $\beta$ is point estimated.  The model optimization problem is non-convex and so a careful initialization strategy is necessary \citep{roberts2016navigating}. \cite{roberts2016navigating}  advocate a deterministic initialization based on the spectral method of moments \citep{arora2013practical} which we refer to below as the spectral initialization.

\subsubsection{Obtaining and using $g$}
In a given experiment we employ the following steps:

\begin{itemize}
\item Create the train-test split
\item In the training set (discovery)
\begin{itemize}
\item explore the documents as desired using STM 
\item choose an estimand (including assigning and validating a label)
\item Identify the mapping function $g$ such that
\begin{align*}
\hat{\boldsymbol{\theta}_i} = g(\boldsymbol{Y}_i, \hat{\boldsymbol{\beta}}, \hat{\boldsymbol{\mu_i}}, \hat{\boldsymbol{\Sigma}})
\end{align*}
\end{itemize}
\item In the test set (evaluation)
\begin{itemize}
\item Using t$g$, obtain our transformed outcome for each document. (see below for details)
\item Estimate treatment effects (using for example the difference of means)
\item Validate model fit and label fidelity in the test set.
\end{itemize}
\end{itemize}

Application of $g$ in STM is equivalent to predicting $\boldsymbol{\theta_i}$ for a held-out document $i$.  This can be accomplished with the recently added \texttt{fitNewDocuments} function in the \texttt{stm} package.  In the STM model, the latent variable $\boldsymbol{\theta_i}$ is a function of a global prior ($\boldsymbol{\mu}, \boldsymbol{\Sigma}$), the topic word parameters $\boldsymbol{\beta}$ and the observed words $\boldsymbol{W_d}$.  The token-level latent variables $\boldsymbol{Z}$ are integrated out.  We have estimated $\boldsymbol{\beta}$ in the train set and in many ways this communicates what the topics substantively contain.  We must also decide how to set our priors $\boldsymbol{\mu}$ and $\boldsymbol{\Sigma}$.  

The \texttt{stm} package offers three options: no prior, the covariate-specific prior and the average prior.  The `no prior' setting sets $\boldsymbol{\mu}$ to a vector of zeroes and $\boldsymbol{\Sigma}$ to be a diagonal matrix with very large diagonals.  The covariate-specific prior uses the observed covariates in the new documents to construct the document-specific prior.  The average prior averages over the values of $\boldsymbol{\mu}$ in the training set and provides a single average prior for  all documents.\footnote{More specifically we take the column means of the $D$ by $K-1$ matrix $\boldsymbol{\mu}$ in the training set which we call $\boldsymbol{\tilde{\mu}}$.  We then recalculate $\boldsymbol{\Sigma}$ as though the update had been made using the new value of $\mu$.  The update is then $\boldsymbol{\tilde{\Sigma}} = \boldsymbol{\Sigma} - \left(\sum_d (\boldsymbol{\eta_d} - \boldsymbol{\mu_d})(\boldsymbol{\eta_d} - \boldsymbol{\mu_d})^T\right) + \left( \sum_d (\boldsymbol{\eta_d} - \tilde{\boldsymbol{\mu}}_d)(\boldsymbol{\eta_d} - \tilde{\boldsymbol{\mu}}_d)^T\right)$.}  

\begin{figure}
\begin{center}
\includegraphics[width=.85\textwidth]{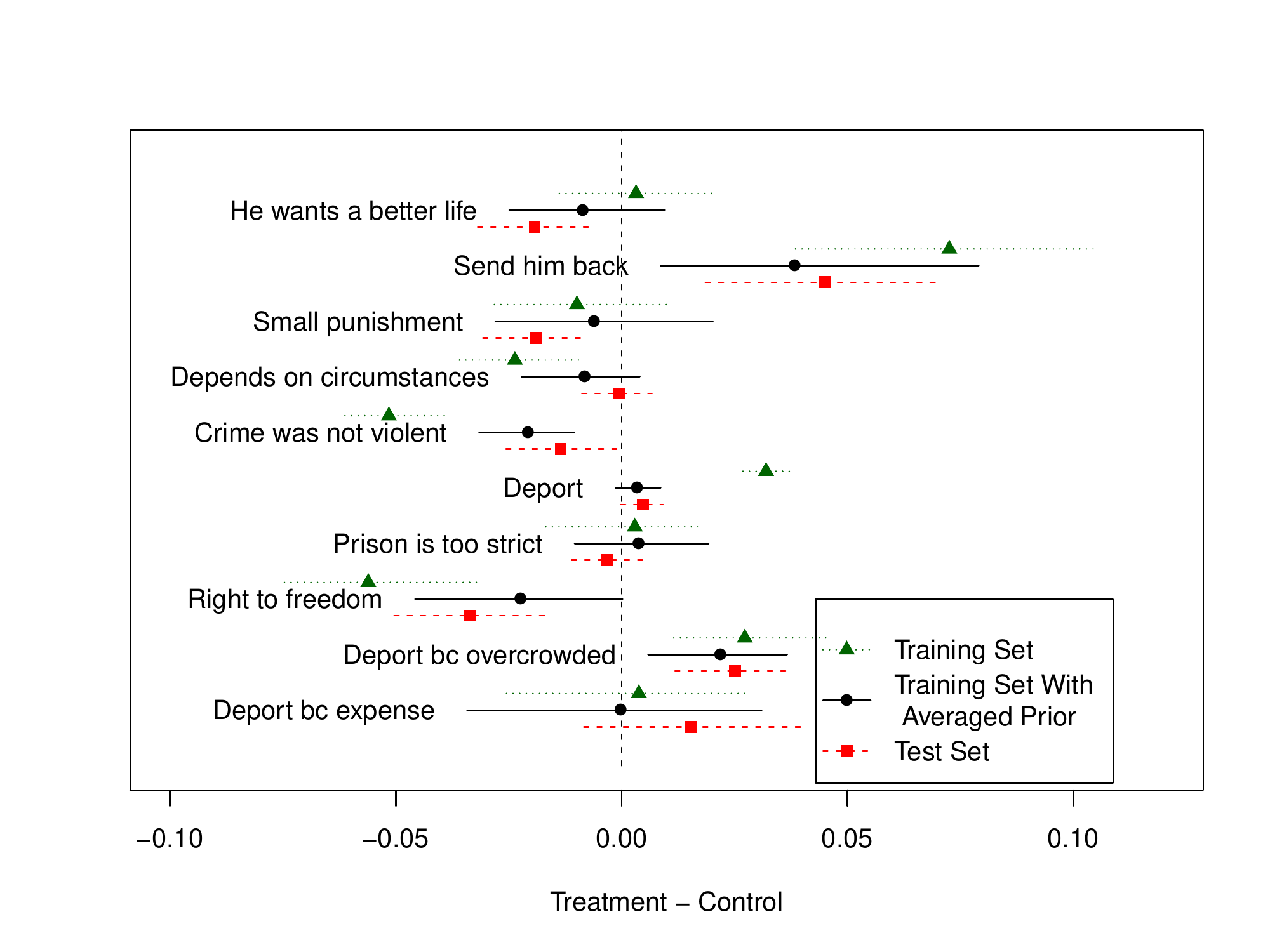}
\end{center}
\caption{Train-Test set effect comparing $g$ using the model estimates  (training set), the training set with averaged prior and the test set.  Note that while the estimates are broadly similar, in general the training set with averaged prior is a closer approximation to what we end up seeing. \label{fig:traintestapp}}
\end{figure}

If we have used only pre-treatment covariates in the STM model we can use any of these strategies.  In our application we do include the treatment and so we cannot use the covariate-specific prior because then the same text would yield two different values of the outcome depending on the treatment assignment.  For our application we use the average prior.  When using a version of $g$ which is not the covariate-specific prior, we recommend that analysts assess effects in the training set using the same procedure as in the test set.  While the effects will generally not be very different (particularly for long documents), maintaining the same procedure should provide a better expectation of test set behavior.  For example, in our application Figure \ref{fig:traintestapp} compares our training set estimates using both the covariate-specific prior and the averaged prior and compares them to the test set (which uses the averaged prior).  Using the average prior to make predictions in the training set before calculating effect estimates gives us a better indication of what we will eventually observe in the test set.

\subsection{Supervised Indian Buffet Process}
\label{sub:sibp}
This appendix provides a brief review of additional estimands and an argument for the use of binary features along with a model for estimating them.

\subsubsection{Additional estimands}
The analyst might also be interested in estimating the effect of an interaction between two components $k$ and $l$.  For example, the researcher might be interested if including military service into a candidate profile has a different effect on candidate ratings if the profile also includes that the candidate is female.  This could be estimated as the Average Component Interaction Effect (ACIE) \citep{hainmueller2013causal} :

\begin{small}
\begin{eqnarray*}
ACIE_{k,l} &=& \int_{Z_{-k,-l}}E[(Y(Z_k=1, Z_l=1,Z_{-k, -l}) - Y(Z_k=1, Z_l=0,Z_{-k, -l})) \\
&-& (Y(Z_k=0, Z_l=1,Z_{-k, -l}) - Y(Z_k=0, Z_l=0, Z_{-k, -l})]m(Z_{-k, -m})dZ_{-k,-m}
\end{eqnarray*}
\end{small}
The ACIE will be the difference between the AMCE for military service for a candidate description that includes information that the candidate is female and the AMCE for military service for a candidate description that does not include this information.

Note that the three complications from the last section also pertain to the case of multidimensional treatments.  If the mapping $g$ between $\boldsymbol{T}$ and $\boldsymbol{Z}$ is not known before defining and reading the treatment texts or the outcome is used in the estimation of these mapping, then an AISV will occur.  Even when using hand coding, researchers should either use a pre-test to determine their coding scheme or use a training/test split to first learn a coding scheme using the responses and then separately estimate the treatment effects.

\subsubsection{The argument for binary features} \label{app:binary}
In this section we explain why we use binary features of texts in order to estimate causal effects.  A different approach to estimating the function $g$ would be to estimate real valued features that explain the text well, such as the principal components of a document term matrix or some other low-dimensional embedding of the observations.  Using these real valued embeddings for $\boldsymbol{Z}$, the impact of $\boldsymbol{Z}$ on $\boldsymbol{Y}$ can be estimated directly.  Using real valued features of documents, however, causes several problems that leads us to use binary features instead.   First, many methods for discovering real valued features incorporate information about the text, but not the response.  For example, we might use the loadings on principal components to describe text-treatments.  This can lead to the discovery of features that explain the content of texts but \textit{do not} explain the response to those texts and therefore are not particularly useful for causal inference. This makes clear that our goal should be to find a low-dimensional representation that explains both the texts and the response well.  Second, using real valued features requires the imposition of a stringent set of functional form assumptions.  This is because even flexibly estimating the response to some continuous feature requires some guidance from a model.  And the more flexible the fit, the more data needed to credibly estimate the response to the continuous treatment.  And as the number of included factors increases, the curse of dimensionality makes it all but impossible to fit anything other than a linear regression.   Alternative approaches, such as an Indian Buffet Process \citep{griffiths2011indian}, yield a binary feature vector about the treatments that are present or absent in a text, but fail to include information about the responses. 

Given the issue with continuous treatments and the importance of including information about the response, we use a method that finds latent features and observation's binary loading on those features, which are then used to estimate treatment effects.  \cite{fong2016discovery} create an unsupervised method for estimating treatments from text data and the responses. They develop a supervised Indian Buffet Process (sIBP) that discovers the topics within the documents that are related to the outcome.  The authors assume that the proportion of documents in each latent feature $k$ is  $\pi_k$, where $\pi_k$ is generated by a stick-breaking algorithm \citep{doshi2009variational}.   Each document can be summarized by treatment vector $Z_j$ where $z_{j,k} \sim$ Bernoulli$(\pi_k)$.  Note that because each individual $z_{j,k}$ is  drawn from a Bernoulli that a treatment document can have more than one latent feature, allowing for multi-dimensional treatments.

The authors assume a mapping from $Z_i$ to the standardized term-document matrix $X_i$ through the D-dimensional vector $A_k$, where $X_i \sim$ MvtNormal$(Z_iA, \sigma_n^2I_D)$.  The latent feature vector $Z_i$ also affects the response $Y_i$ through the normal, $Y_i \sim$ Normal$(Z_i\beta, \tau^-1)$ where $\tau \sim$ Gamma$(a,b)$.  Thus with the model the authors both want to discover the latent treatments $Z_i$ \emph{and} estimate their influence on the outcome by estimating $\beta$.  The authors use variational approximation to estimate these parameters.

\cite{fong2016discovery} apply the sIBP to the training data in order to learn $g$.  In the test set \cite{fong2016discovery} use $g$ to infer the treatments that are present in a particular text, but alter the inference to avoid conditioning on the dependent variable.  They do this because otherwise the inferred treatments present in the test set will depend upon the observation's response to that text, which creates obvious problems for causal inference.

Once the latent treatments are inferred in the test set documents, their effect can be estimated using any procedure that might be used to analyze an experiment.  \cite{fong2016discovery} use a simple linear regression with each of the latent features as the regressors to estimate the effects of the treatments.  More complicated models could be used to estimate interactions or to extrapolate effects to a different population of documents.

\subsection{Stability Across Train-Test Splits}
\label{app:stability}
Our approach does not require stability of analysis across different train-test splits.  Different train-test splits might lead to different discovery phases which in turn yield different estimands and test sets where we can evaluate that estimand.  Nevertheless, we might be slightly uncomfortable with the idea that the particular randomization into the train-test spit yield quite different estimands (and papers) at the end of the process.  As such we wanted to evaluate the stability of the STM under different samples of a fixed population. 

In a formal sense we are interested in studying the posterior contraction rates of the model, a problem taken up analytically in \cite{tang2014understanding} for the related Latent Dirichlet Allocation model.  However, we are far more interested in understanding performance in practice and whether different train-test splits lead to substantively different topic-word distributions ($\boldsymbol{\beta}$), different document-topic proportions ($\boldsymbol{\theta}$) and different covariate effects.  As the number of documents increase or the topics are more sharply defined stability will improve.  For this demonstration we use the Poliblog data \citep{poliblog}, a collection of around 13,246 posts from six different political blogs in the runup to the 2008 American presidential election.  We use this because it is readily available for use with the \texttt{stm} package and is roughly representative of the document lengths that we often see in \texttt{stm} applications overall.  We would expect that the diversity of topics in political blogs would make the problem harder than the more focused open-ended response case, but the length of the documents would make it easier.  

We started by estimating the model on the full set of documents with 20 topics using the spectral initialization.  We consider this to be the ``truth'' because the unattainable stability ideal would be that the subsamples provide the same answer as the full set of documents.  We then choose two prominent topics to be our ``outcomes'' a topic about Obama and a topic about War (particularly Iraq and Afghanistan).  In each simulation we choose the topic that most closely approximates our two chosen outcomes, emphasizing that the labels 'Obama' and 'War' may well not be good approximations for the topic in the subsample.

Because of the multimodality problem in topic models, instability could arise from two sources: differences in the local mode discovered and differences in the data observed.  We investigate this by considering three different initialization strategies:
\begin{itemize}
\item[1)] Cold Spectral Start \\
Using the spectral initialization on the subsample.  This is reflective of current practice.
\item[2)] Warm Spectral Start \\
Use the complete data to initialize the model.  This would create an analyst-induced SUTVA violation as it shares information from the test set.  However, it is suggestive of what might be achievable by providing more stable intializations.
\item[3)] Warm Oracle Start \\
Use the results of the \textit{converged} model on the full sample to initialize each subsample.  This is an infeasible estimator.  The instability in this estimate cannot be reduced by a better initialization strategy.
\end{itemize}

\begin{figure}[!h]
\caption{Stability of $\theta$ in Simulations of Train-Test Splits on Real Data. \label{fig:theta}}
  \centering
  \begin{minipage}{0.45\hsize}
    \begin{center}
      {\small {\bf Sample=5000 with Cold Spectral start}}
      \includegraphics[height=4cm]{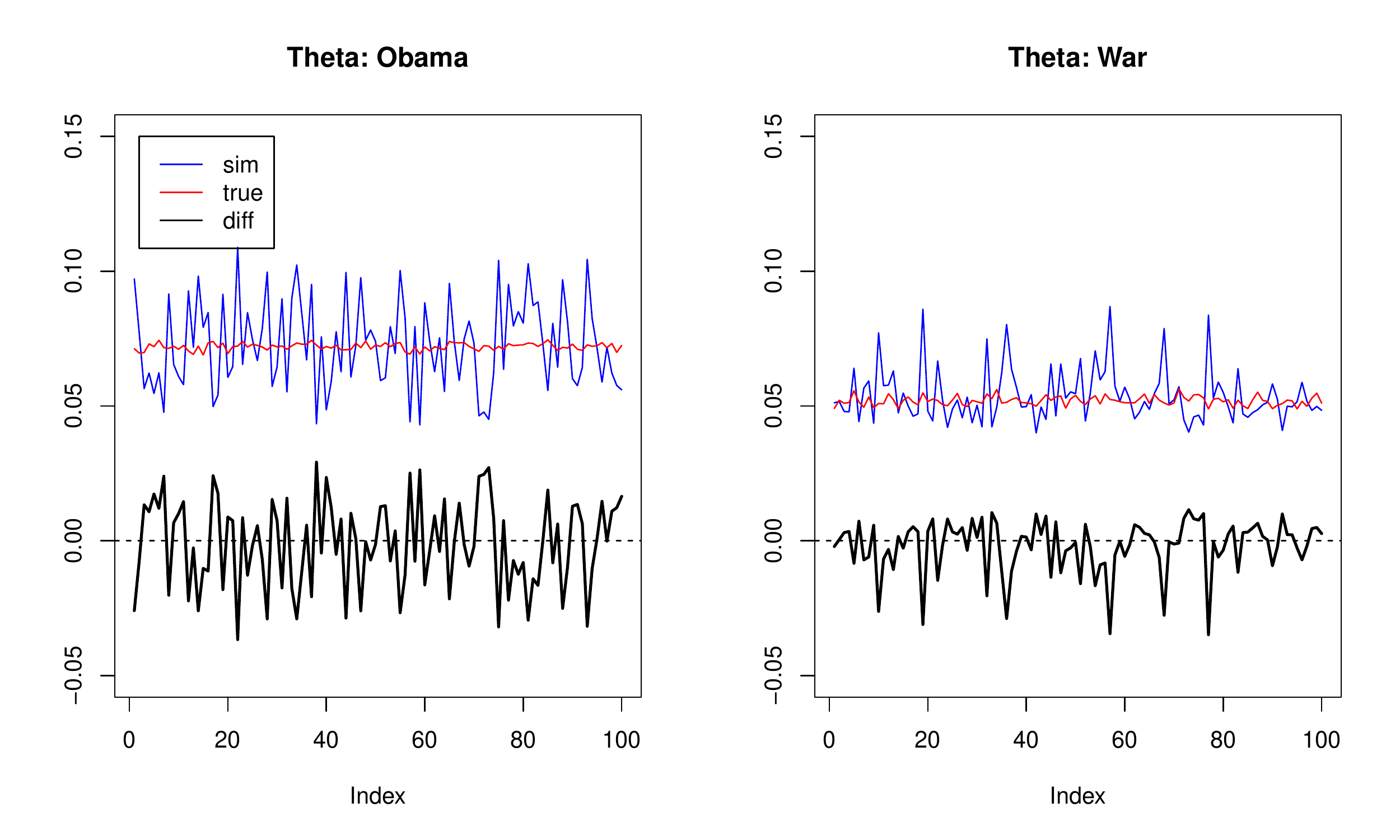}
    \end{center}
  \end{minipage}
  \hspace{0.3in}
  \begin{minipage}{0.45\hsize}
    \begin{center}
      {\small {\bf Sample=1000 with Cold Spectral Start}}
      \includegraphics[height=4cm]{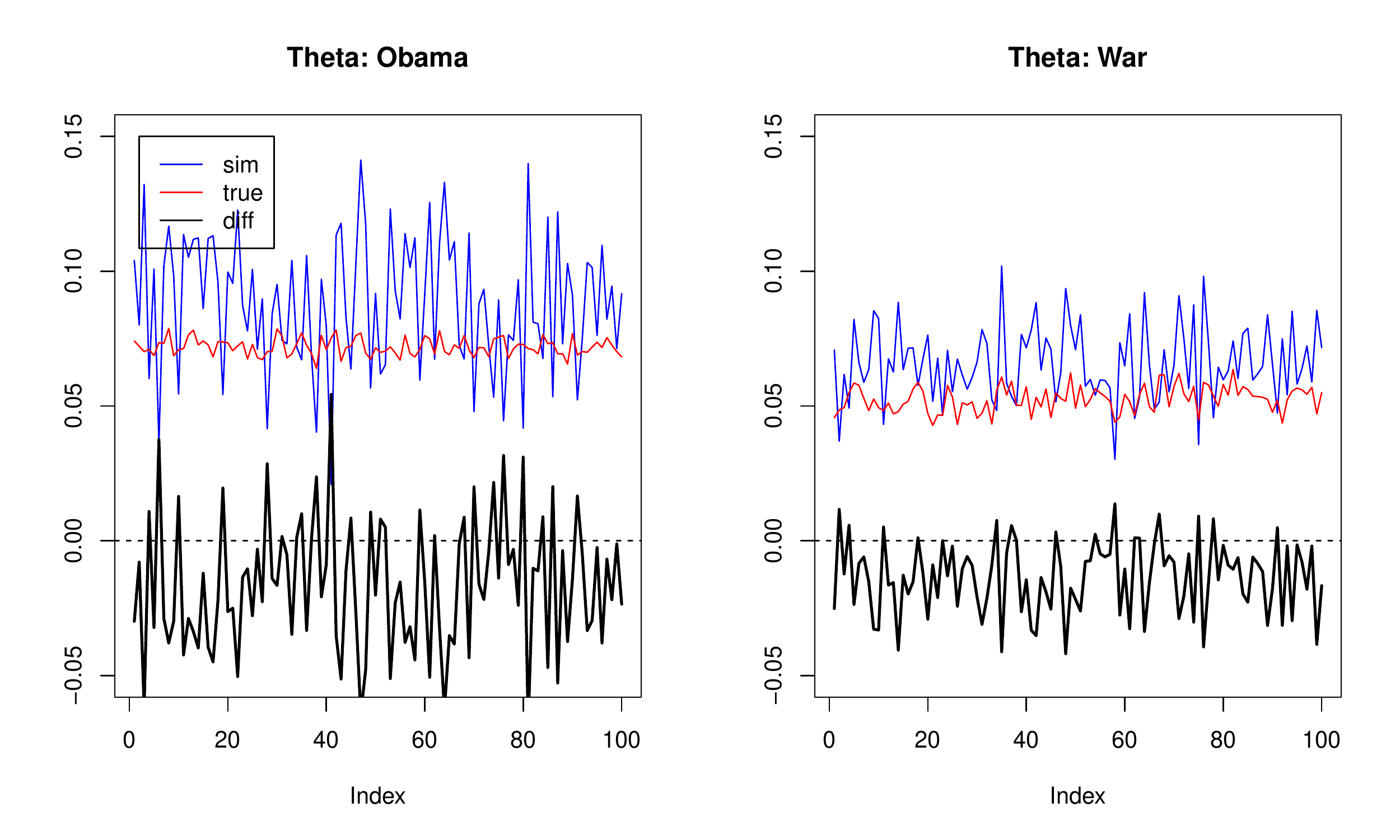}
    \end{center}
  \end{minipage}
  
  \vspace{0.2in}
  
  \begin{minipage}{0.45\hsize}
    \begin{center}
    {\small {\bf Sample=5000 with Warm Spectral Start}}
    \includegraphics[height=4cm]{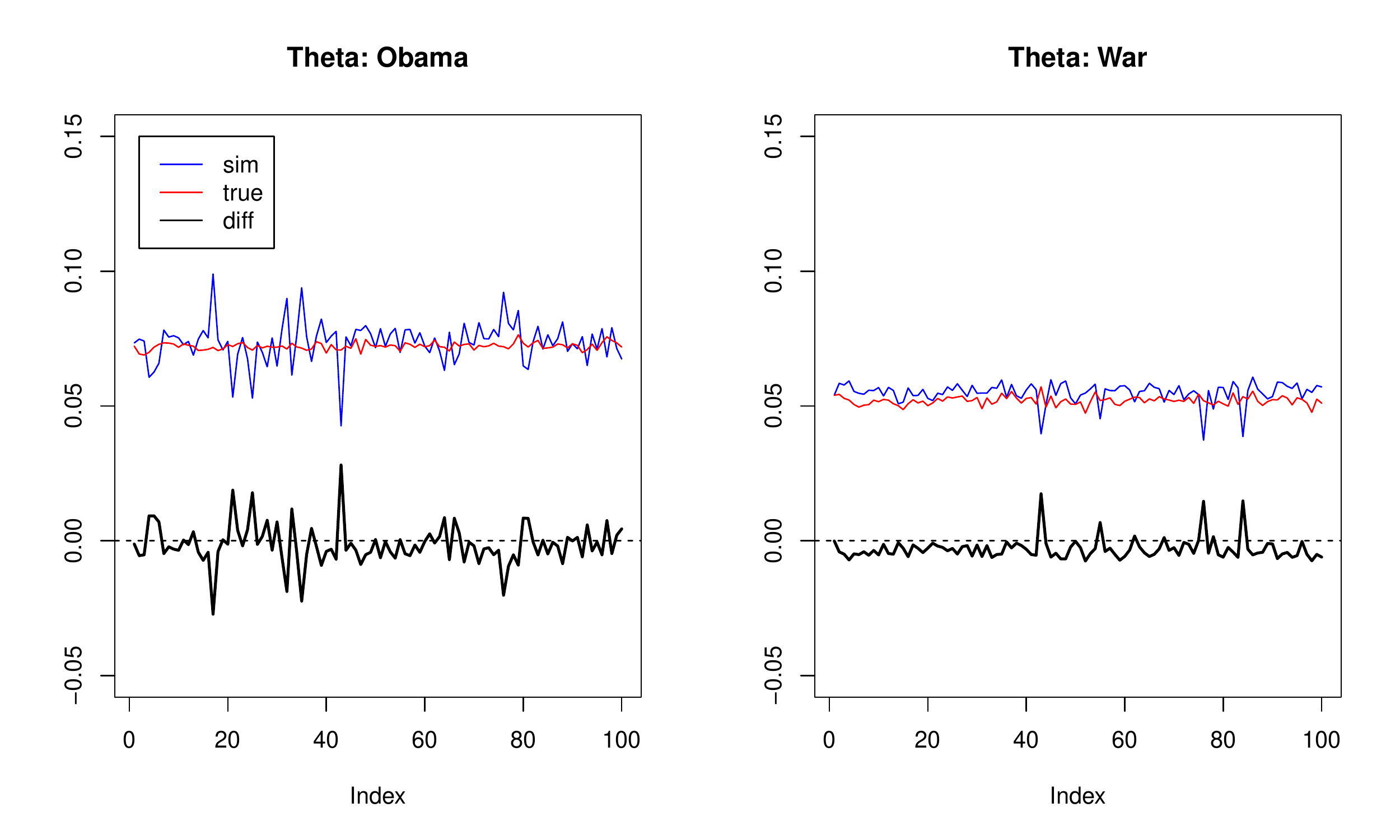}
  \end{center}
\end{minipage}
\hspace{0.3in}
\begin{minipage}{0.45\hsize}
  \begin{center}
    {\small {\bf Sample=1000 with Warm Spectral Start}}
    \includegraphics[height=4cm]{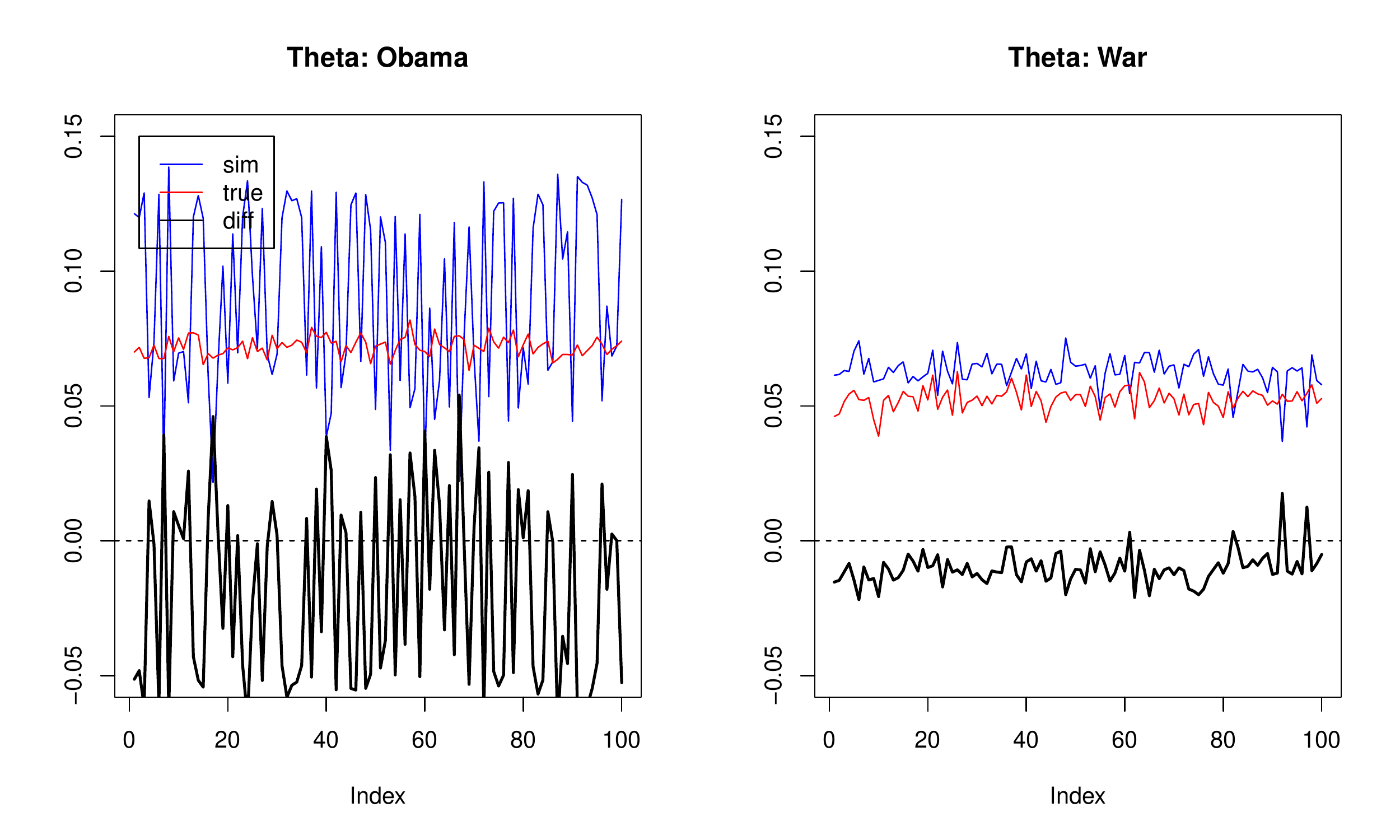}
  \end{center}
\end{minipage}
\vspace{0.2in}

\begin{minipage}{0.45\hsize}
  \begin{center}
    {\small {\bf Sample=5000 with Warm Oracle Start}}
    \includegraphics[height=4cm]{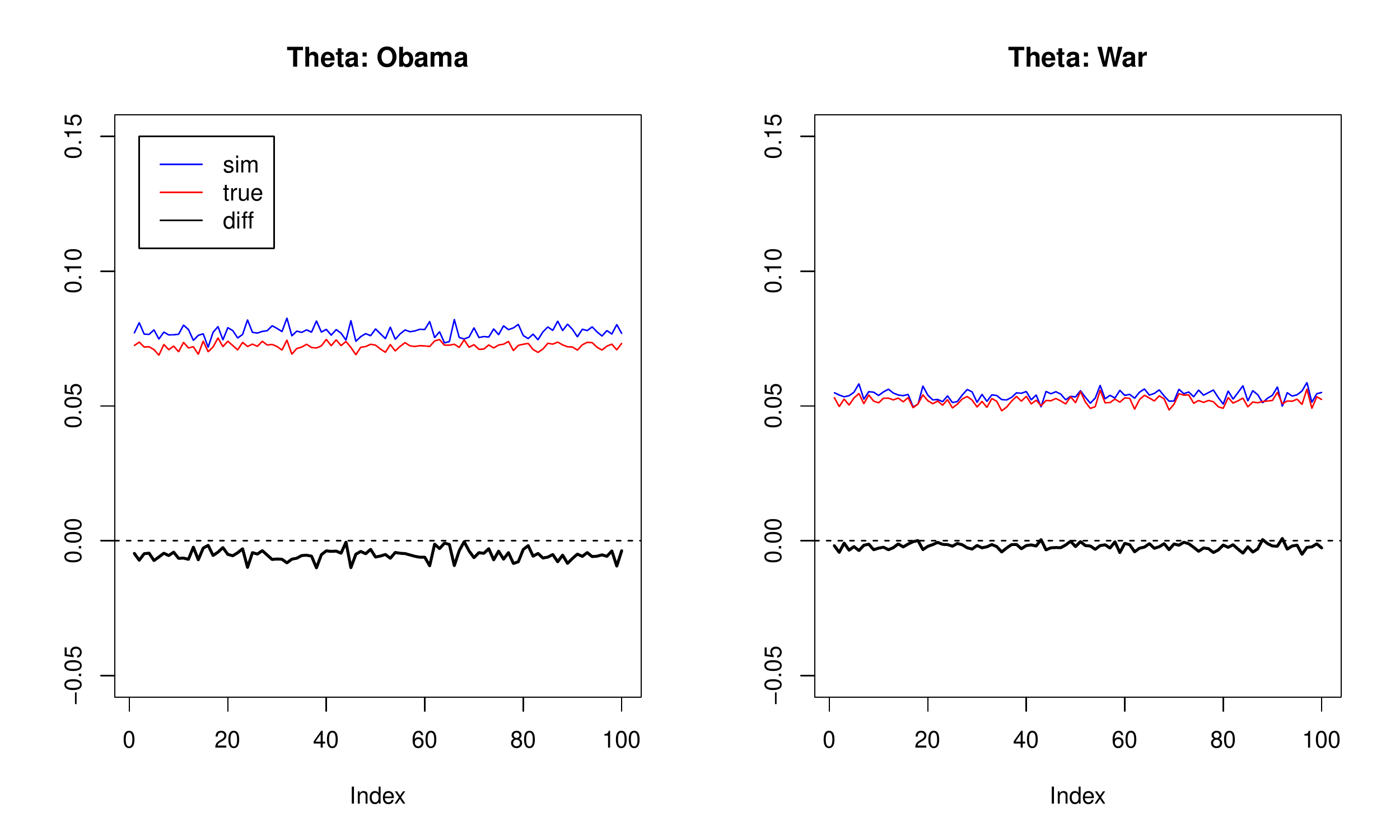}
  \end{center}
  \end{minipage}
  \hspace{0.3in}
  \begin{minipage}{0.45\hsize}
    \begin{center}
      {\small {\bf Sample=1000 with Warm Oracle Start}}
      \includegraphics[height=4cm]{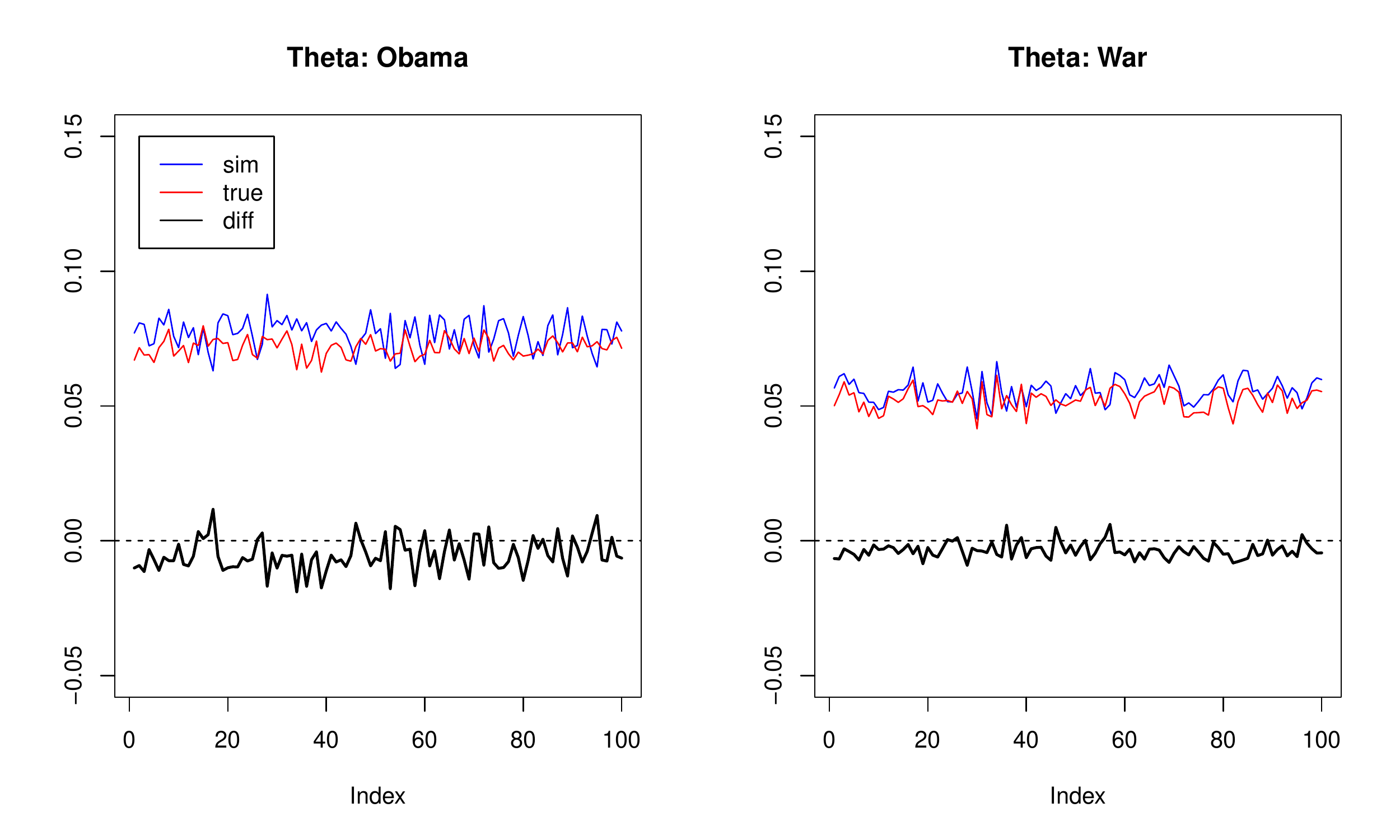}
    \end{center}
  \end{minipage}
\end{figure}

In each case we run the model on the sample sizes 100 times and plot the results along with the `truth' as defined by the full document set.  Figure \ref{fig:theta} shows the results for the average proportion of the topic use in the corpus.

As we can see the results are reasonably stable at 5000 documents for a corpus of this complexity and less so with 1000.  The warm spectral start shows considerable improvement for the 5000 document case suggesting that at least at this scale, we might see substantial gains from an initialization specificly designed to be more stable across splits.  The near perfect stability of the warm oracle start  for the 5000 document case suggests it is a matter of the initialization and not necessarily the data itself, where for 1000 documents there is evidence that some level of the instability is unavoidable given the model.

\begin{figure}[!h]
\caption{Stability of $\beta$ in Simulations of Train-Test Splits on Real Data. \label{fig:beta}}

  \centering
  \begin{minipage}{0.45\hsize}
  \begin{center}
    {\small {\bf Sample=5000 with Cold Spectral Start}}
    \includegraphics[height=4cm]{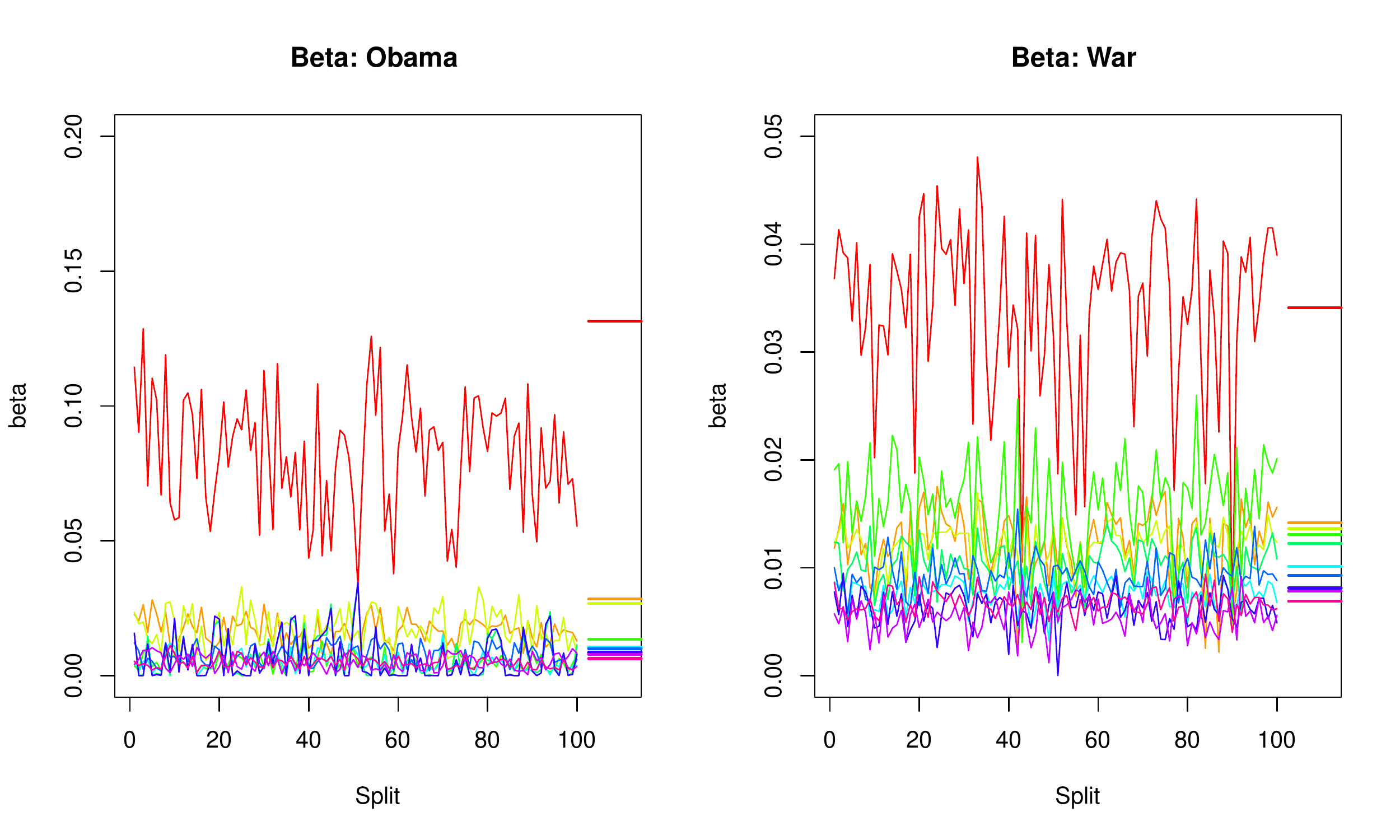}
  \end{center}
  \end{minipage}
  \hspace{0.3in}
  \begin{minipage}{0.45\hsize}
  \begin{center}
    {\small {\bf Sample=1000 with Cold Spectral Start}}
    \includegraphics[height=4cm]{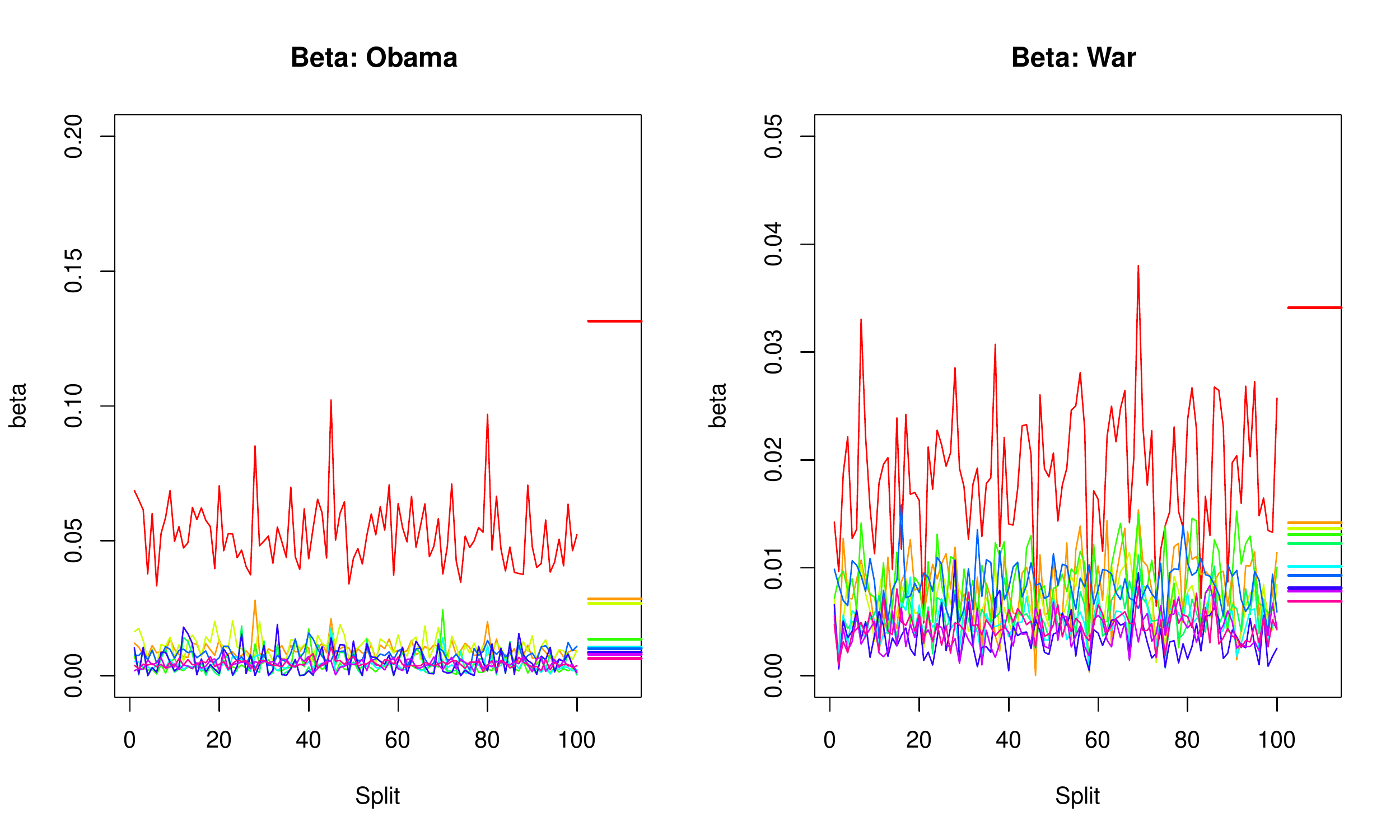}
  \end{center}
  \end{minipage}

\vspace{0.2in}
  \begin{minipage}{0.45\hsize}
  \begin{center}
    {\small {\bf Sample=5000 with Warm Spectral Start}}
    \includegraphics[height=4cm]{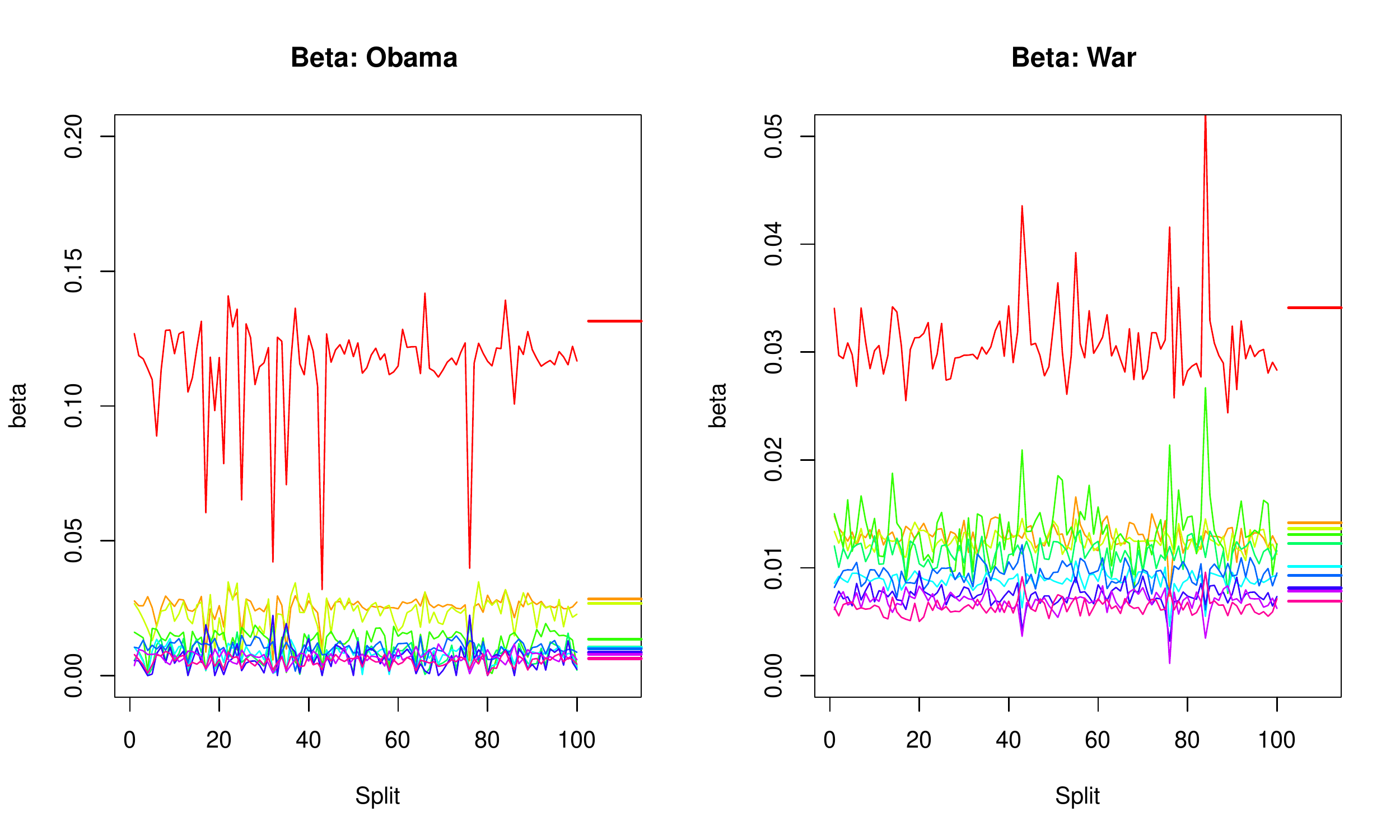}
  \end{center}
  \end{minipage}
  \hspace{0.3in}
  \begin{minipage}{0.45\hsize}
  \begin{center}
    {\small {\bf Sample=1000 with Warm Spectral Start}}
    \includegraphics[height=4cm]{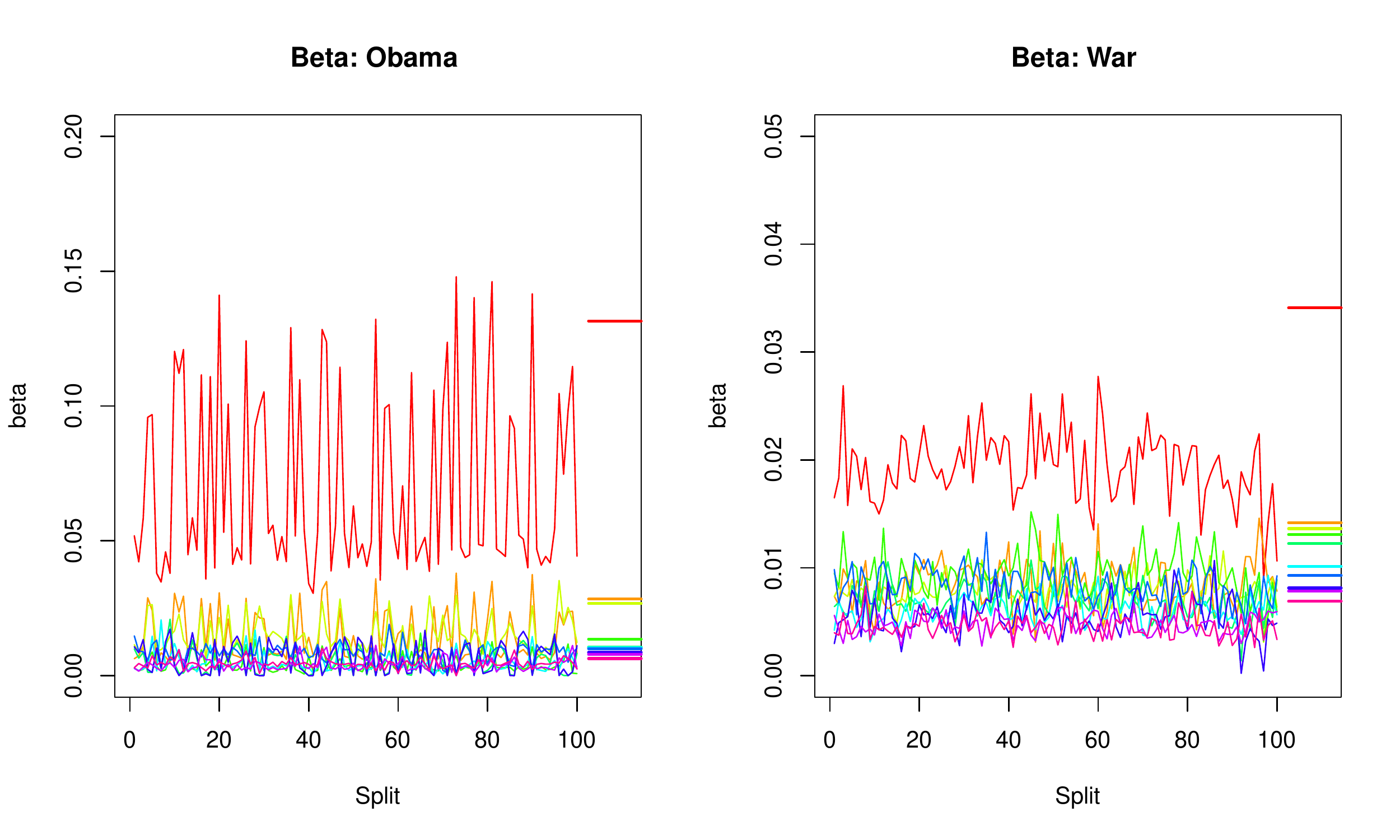}
  \end{center}
  \end{minipage}

  \vspace{0.2in}
  \begin{minipage}{0.45\hsize}
  \begin{center}
    {\small {\bf Sample=5000 with Warm Oracle Start}}
    \includegraphics[height=4cm]{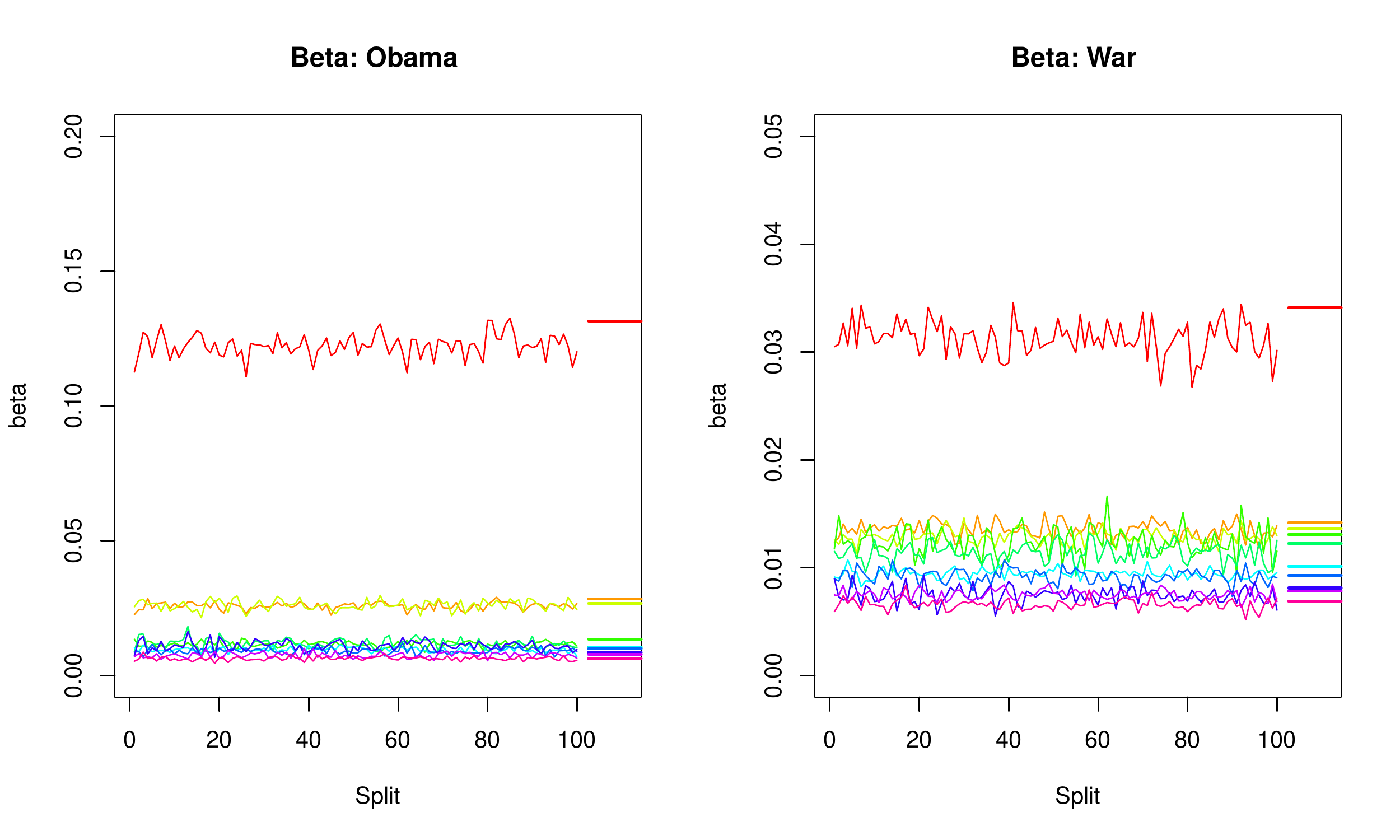}
  \end{center}
  \end{minipage}
  \hspace{0.3in}
  \begin{minipage}{0.45\hsize}
  \begin{center}
    {\small {\bf Sample=1000 with Warm Oracle Start}}
    \includegraphics[height=4cm]{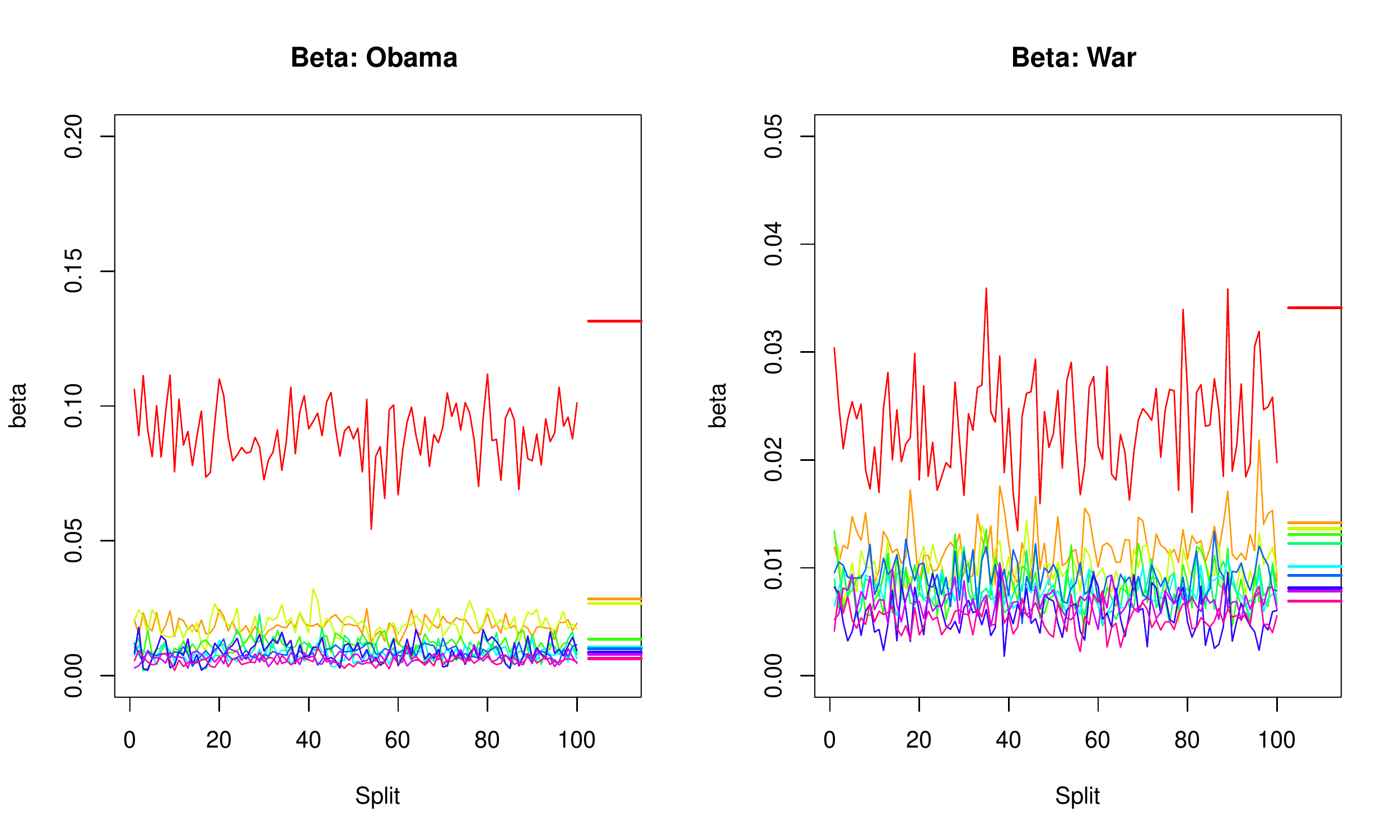}
  \end{center}
  \end{minipage}
\end{figure}

We can also examine the word-distributions themselves.  Figure \ref{fig:beta} shows the proportion of mass associated with each of the top ten words in the topic (as chosen by the full model).  The horizontal tickmarks on the right show the estimate in the full data.

Generally speaking the models correctly preserve the rank ordering of the most prominent words in each topic, but the estimates can often be substantially incorrect.  We do emphasize that there is relatively little information with which to estimate these parameters and so we would expect to see more instability than in the simulations for $\theta$.

\begin{figure}[!h]
\caption{Stability of Covariate Effect in Simulations of Train-Test Splits on Real Data. \label{fig:tau}}
  \centering
  \begin{minipage}{0.45\hsize}
  \begin{center}
    {\small {\bf Sample=5000 with Cold Spectral Start}}
    \includegraphics[height=4cm]{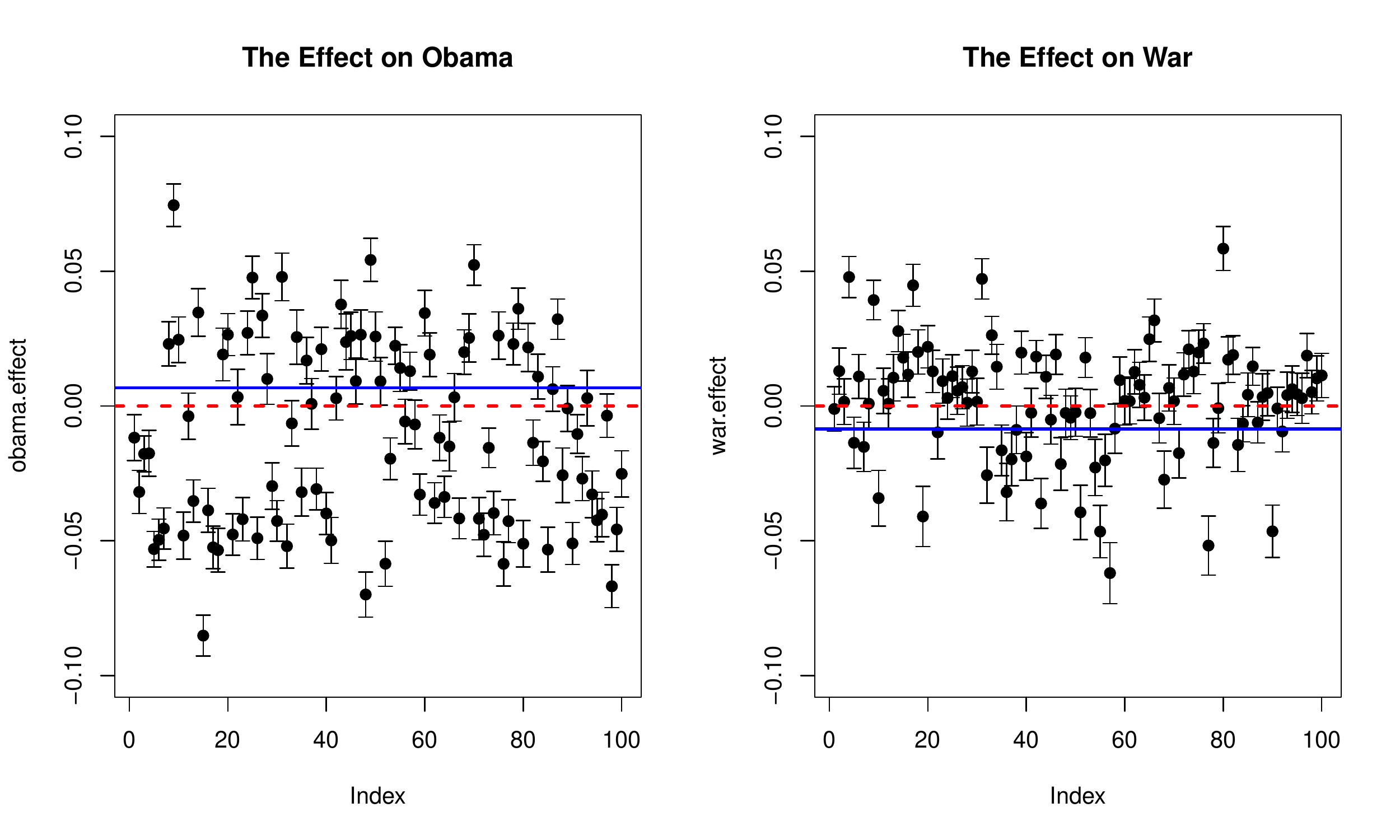}
  \end{center}
  \end{minipage}
  \hspace{0.3in}
  \begin{minipage}{0.45\hsize}
  \begin{center}
    {\small {\bf Sample=1000 with Cold Spectral Start}}
    \includegraphics[height=4cm]{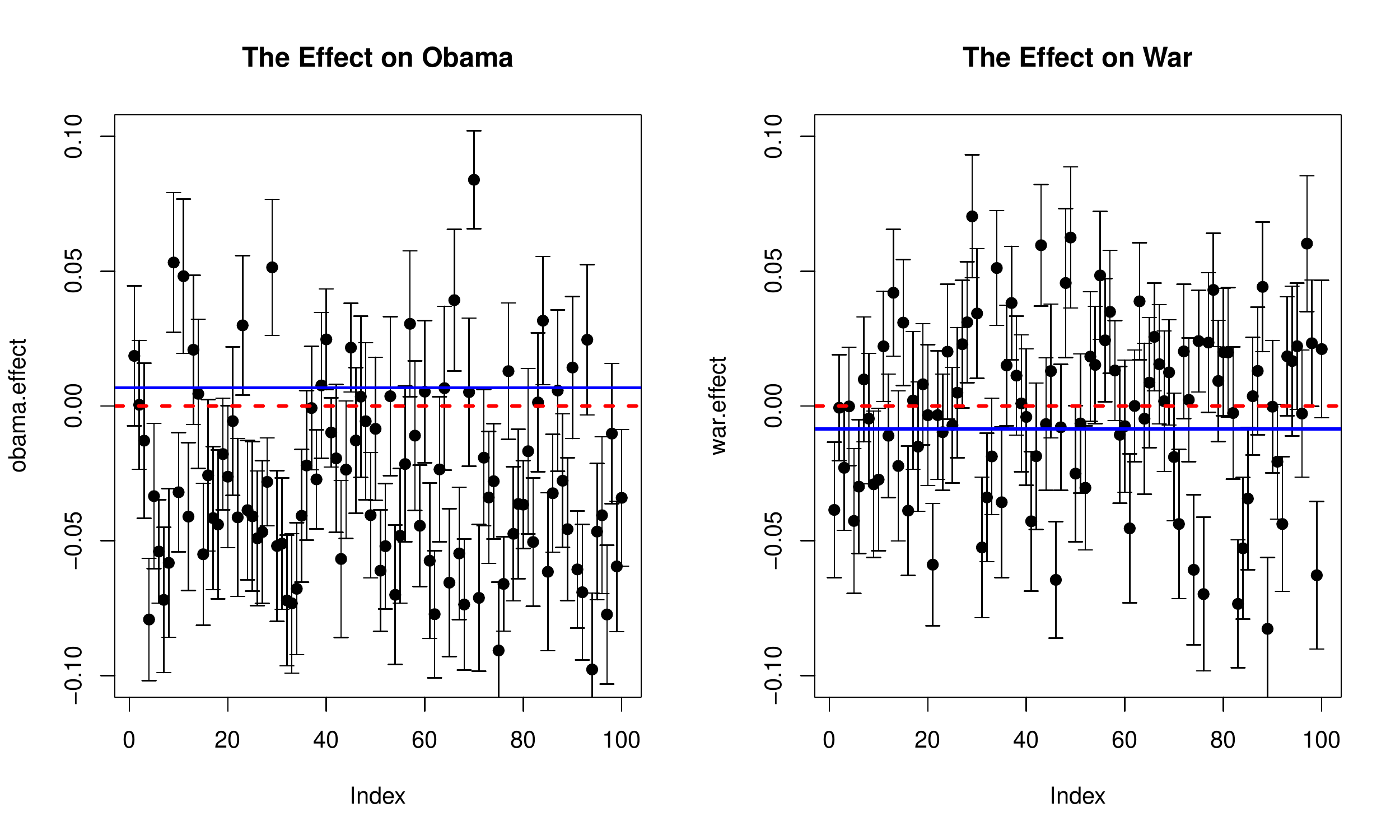}
  \end{center}
  \end{minipage}

\vspace{0.5in}
  \begin{minipage}{0.45\hsize}
  \begin{center}
    {\small {\bf Sample=5000 with Warm Spectral Start}}
    \includegraphics[height=4cm]{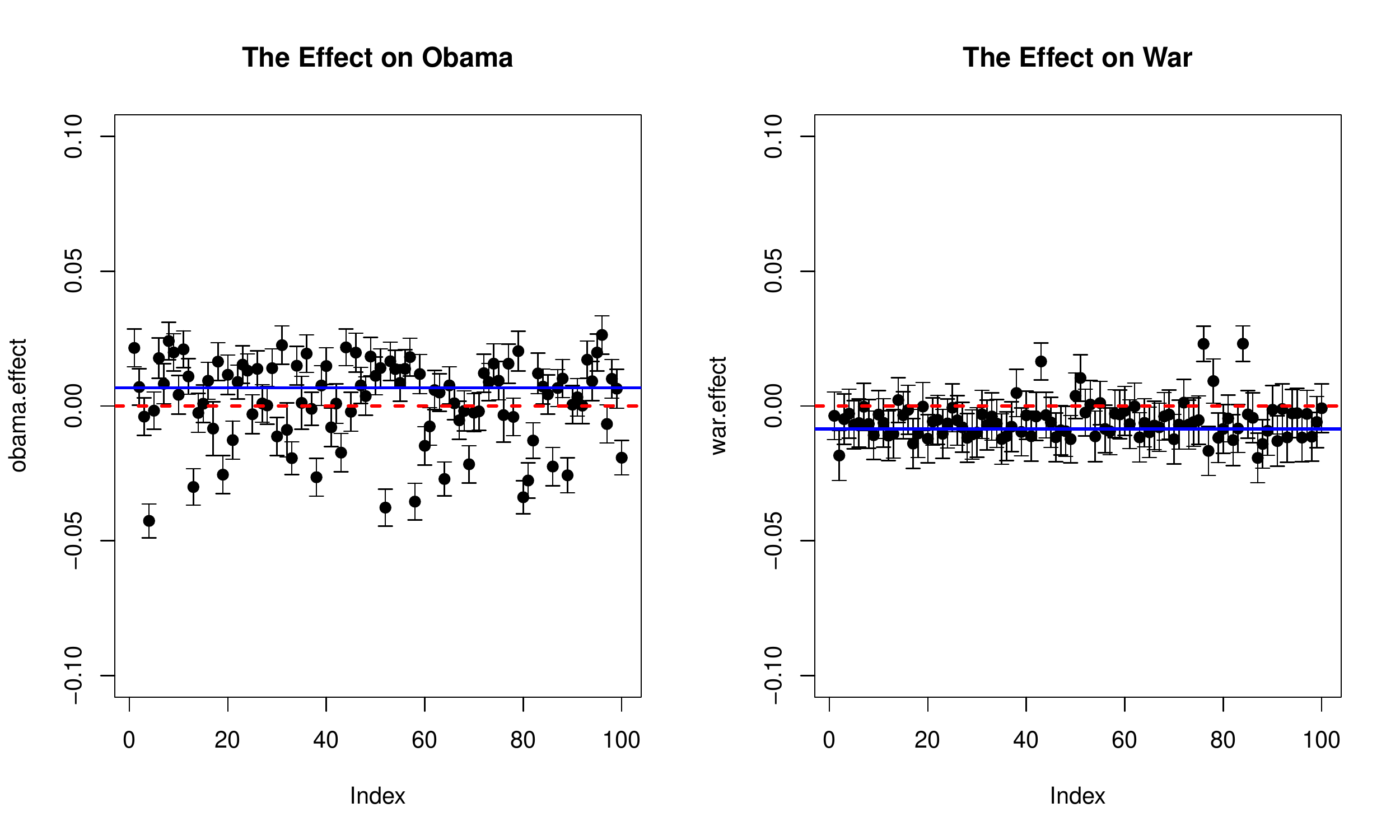}
  \end{center}
  \end{minipage}
  \hspace{0.3in}
  \begin{minipage}{0.45\hsize}
  \begin{center}
    {\small {\bf Sample=1000 with Warm Spectral Start}}
    \includegraphics[height=4cm]{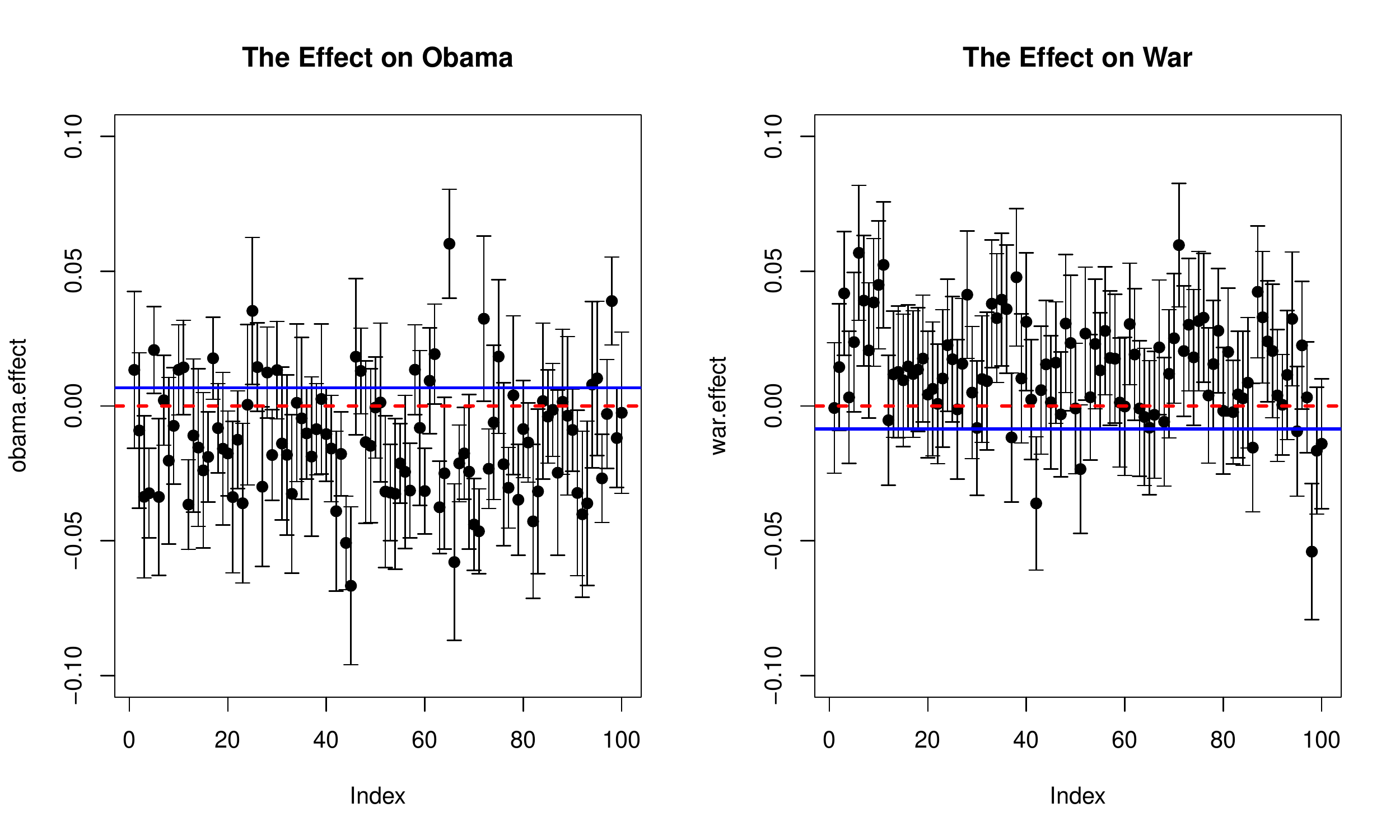}
  \end{center}
  \end{minipage}

  \vspace{0.5in}
  \begin{minipage}{0.45\hsize}
  \begin{center}
    {\small {\bf Sample=5000 with Warm Oracle Start}}
    \includegraphics[height=4cm]{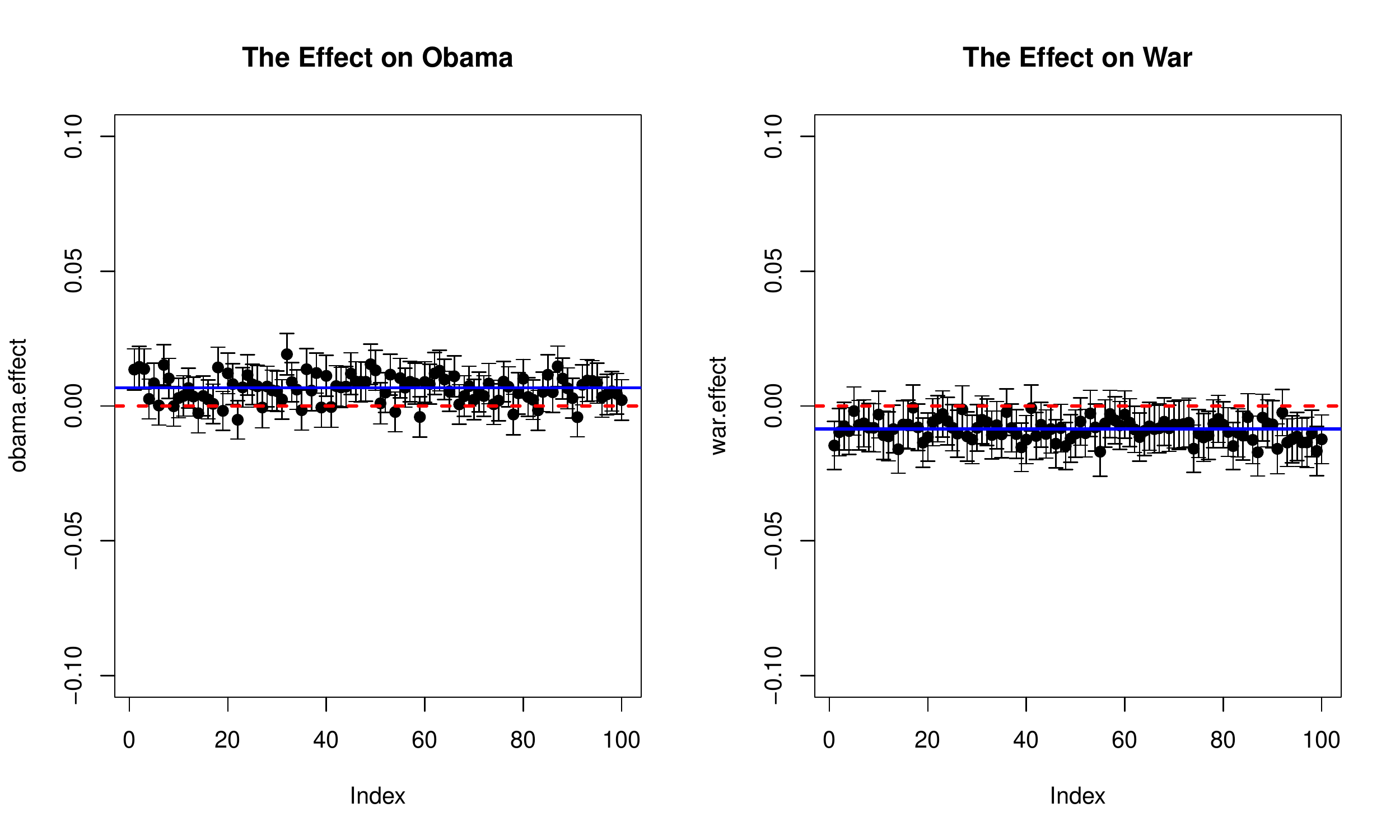}
  \end{center}
  \end{minipage}
  \hspace{0.3in}
  \begin{minipage}{0.45\hsize}
  \begin{center}
    {\small {\bf Sample=1000 with Warm Oracle Start}}
    \includegraphics[height=4cm]{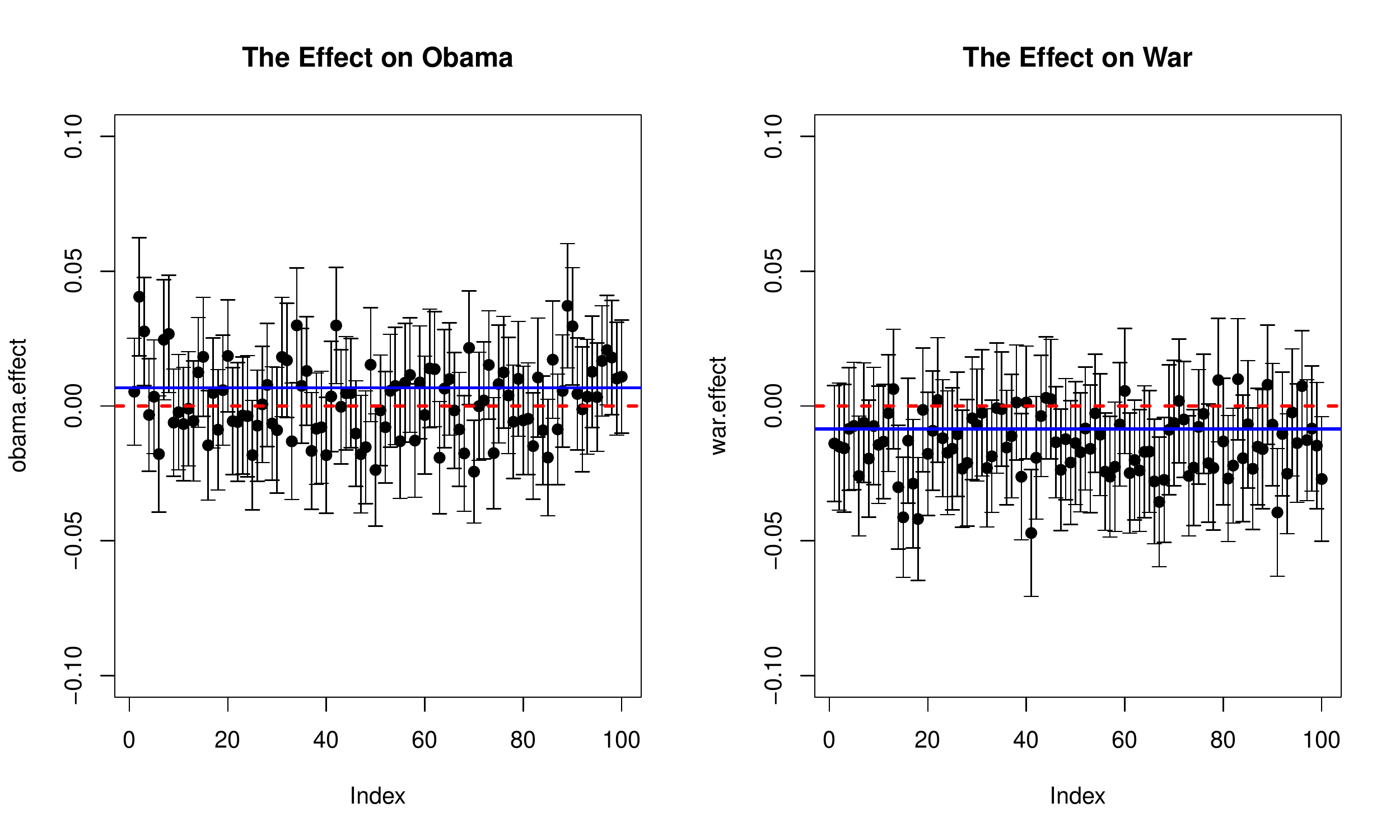}
  \end{center}
  \end{minipage}
\end{figure}

Finally we present estimates of ``treatment effects.''  Here we use the binary rating variable (indicating whether the blog is liberal or conservative) as a treatment.  This is clearly not randomly assigned and we use it simply because it is a binary covariate we would expect to influence the outcome in some way.  We plot the estimate with a $95\%$ confidence interval in Figure \ref{fig:tau} along with the estimate in the complete dataset shown in blue.

Once again we can see some substantial variability that appears to be reducible via a more stable initializations.  We emphasize that we should not expect these confidence intervals to have proper coverage as in every case the estimand is different.  Indeed the high variance on the Obama topic of the warm spectral start is a good indication that the estimand is changing substantially in each different split.  What the relatively tight set of estimates in the two warm starts suggest is that this might be avoidable with a different initialization.

In summary, there is a significant degree of instability across splits.  Again, this is not a problem in a technical sense as $g$applied to the test set will still provide a proper estimator of that specific estimand.  What these simulations do suggest is that further research into more stable initialization strategies might substantially reduce the amount of instability across train-test splits.  

There are several limitations to this simulation study: most notably that we neither know the actual truth nor can we be sure what the scope conditions are for these results to apply to other datasets.  We also cannot simulate the stability of the entire discovery process, only that a particular model is comparable across subsamples.  Hoping for stability in discovery may be quixotic as the very idea of discovery itself may imply some level of instability.  
\clearpage

\subsection{Full set of experimental results for the Immigration Experiment}
\label{app:experiment}
\begin{table}[ht!]
\centering
\begin{tabular}{|p{.55in}|p{1.5in}|p{3.5in}|}
  \hline
 & Label & Highest Probability Words \\
  \hline
Topic 1 & He wants a better life & didnt, want, pay, better, life, probabl, isnt \\
  Topic 2 & Send him back & back, countri, send, home, well, charg, troubl \\
  Topic 3 & Small punishment & offens, reason, like, chanc, first, can, citizen \\
  Topic 4 & Depends on circumstances & come, depend, doesnt, free, feel, law, shouldnt \\
  Topic 5 & Crime was not violent & crime, commit, violent, immigr, wasnt, look, never \\
  Topic 6 & Deport & deport, that, give, counti, peopl, look, guilti \\
  Topic 7 & Prison is too strict & enter, anyth, right, live, realli, illeg, anybodi \\
  Topic 8 & Right to freedom & just, tri, get, hes, came, freedom, put \\
  Topic 9 & Deport bc overcrowded & sent, prison, think, alreadi, anoth, done, hasnt \\
  Topic 10 & Deport bc expense & dont, think, know, time, need, serv, crimin \\
   \hline
\end{tabular}
\caption{\label{topicwords1} Experiment 1, Words most representative of topics.}
\end{table}

\begin{table}[ht!]
\centering
\begin{tabular}{|p{.55in}|p{1.5in}|p{3.5in}|}
  \hline
 & Label & Representative Document \\
  \hline
Topic 1 & He wants a better life & we're the land of opportunity everybody wants a better life                                                                                                                                                                                                                                                                                                                                                                                                                                                                                                                                                                                               \\
  Topic 2 & Send him back & send him back to his country                                                                                                                                                                                                                                                                                                                                                                                                                                                                                                                                                                                                                              \\
  Topic 3 & Small punishment & "it was his first offense, didn't hurt anybody, maybe a fine though, probation or something. that's nice serious like murder or robbery"                                                                                                                                                                                                                                                                                                                                                                                                                                                                                                                  \\
  Topic 4 & Depends on circumstances & it depends on reaason why he is coming into state if he was coming to beter himself its ok if he has a record he should be disbarred or deported                                                                                                                                                                                                                                                                                                                                                                                                                                                                                                          \\
  Topic 5 & Crime was not violent & because he didnt commit a crime that was effecting someone else's individual liberties                                                                                                                                                                                                                                                                                                                                                                                                                                                                                                                                                                    \\
  Topic 6 & Deport & he should be deported                                                                                                                                                                                                                                                                                                                                                                                                                                                                                                                                                                                                                                     \\
  Topic 7 & Prison is too strict & because he didnt do anything except illegally enter                                                                                                                                                                                                                                                                                                                                                                                                                                                                                                                                                                                                       \\
  Topic 8 & Right to freedom & Because he's just trying to get his freedom. Maybe he's trying to away from a tough situation/that country-maybe it's not good for him.                                                                                                                                                                                                                                                                                                                                                                                                                                                                                                                   \\
  Topic 9 & Deport bc overcrowded & he should be sent to prison in another country our prisons are over crowded already                                                                                                                                                                                                                                                                                                                                                                                                                                                                                                                                                                       \\
  Topic 10 & Deport bc expense & because i think he shold be deported-p-i don't think he should be supported in our prison system and i don't think he should be allowed to immigrate here                                                                                                                                                                                                                                                                                                                                                                                                                                                                                                 \\
   \hline
\end{tabular}
\caption{\label{topicdocs} Representative documents of each topic.}
\end{table}

\begin{table}[ht]
\centering
\begin{scriptsize}
\begin{tabular}{|p{.55in}|p{1.5in}|p{3.5in}|}
  \hline
 & Label & Highest Probability Words \\
  \hline
Topic 1 & Prison, committed crimes & crime, commit, violent, illeg, immigr, punish, convict \\
  Topic 2 & Prison, repeat offender & state, unit, offend, need, offens, enter, repeat \\
  Topic 3 & Prison, because of the law & law, jail, time, alreadi, come, will, prior \\
  Topic 4 & Prison, but depends on circumstances & serv, prison, sentenc, one, time, know, feel \\
  Topic 5 & Looking for better life & person, crimin, govern, better, life, good, stay \\
  Topic 6 & No prison bc taxpayers & imprison, money, believ, allow, origin, tax, taxpay \\
  Topic 7 & Stay, but depends on circumstances & think, illeg, enter, dont, peopl, just, man \\
  Topic 8 & Deport, if needed prison & countri, deport, prison, back, sent, home, send \\
   \hline
\end{tabular}
\caption{\label{topicwords2} Experiment 2, Words most representative of topics.}
\end{scriptsize}
\end{table}

\begin{table}[ht]
\centering
\begin{scriptsize}
\begin{tabular}{|p{.55in}|p{1.5in}|p{3.5in}|}
  \hline
 & Label & Representative Document \\
  \hline
Topic 1 & Prison, committed crimes & He committed crimes, and most importantly, violent crimes, so should be convicted on that.  I am not concerned as much with his immigration status, although the fact that he keeps returning after deportation should be taken into account.  I am not judging his origin, just his crimes. \\
  Topic 2 & Prison, repeat offender & Because the man  is a citizen of another country. that is not main matter. illegaly entered the united states this is main mistake. so that man is lock to prison \\
  Topic 3 & Prison, because of the law & It'll be the first lesson for him to obey the laws. Secondly, teach him to think before do the things.  \\
  Topic 4 & Prison, but depends on circumstances & I really don't know how to judge the severity of this crime and what an appropriate punishment would be.  If someone were to sneek into a club, ball game, or onto private property in general they would probably be at least subject to trespassing charges but my guess is this would involve only a fine and not prison time.  However, this crime is probably more severe.  If someone were to sneek onto the grounds of the White House I think they would likely be charged with crimes that involve prison time however illegally entering the United States might be considered less sever than such a crime.  I guess I'd have to here arguments for and against before I could come to some sort of conclusion.  In general, I don't have a strong sense of the harms of the crime. \\
  Topic 5 & Looking for better life & He has never been in trouble before, he is obviously looking for safety and he should be helped along the road to becoming a citizen. \\
  Topic 6 & No prison bc taxpayers & His entry into the US costed the US tax payer nothing. His crime and imprisonment would cost the US tax payer thousands of dollars in food, shelter, health care, etc... for this man. It is cheaper to instead remove the individual from the US at his own expense, if possible.  \\
  Topic 7 & Stay, but depends on circumstances & I think it depends on what country he came from, and why he entered the US illegally.  If he's a refugee that was no longer safe waiting for the US to approve his arrival and is requesting citizenship, it's quite a different case than if he had never bothered to try entering legally, just came because he wanted to make money, and was planning on staying here illegally without ever becoming a citizen.  The government should find out more of the reasons why this man entered illegally and then base the punishment on that. \\
  Topic 8 & Deport, if needed prison & The man should be held in prison before being deported. His home country should take him and if they don't then he should be held in prison. This man shouldn't be in the country and should leave as soon as possible. \\
   \hline
\end{tabular}
\caption{\label{topicwords3} Experiment 2, Responses documents of each topic.}
\end{scriptsize}
\end{table}

\begin{table}[ht!!!]
\centering
  \begin{scriptsize}
\begin{tabular}{r p{.2\textwidth} p{.7\textwidth}}
  \hline

 & Label & Representative Document \\ 
  \hline
Topic 1 & Limited punishment with help to stay in country, complaints about immigration system & with all of the ""exceptional america"", ""anyone can get rich"" propaganda this country throws out(not exactly the truth since we are no longer exceptional(literacy, happiness, health care), and the fact some people are actually taking us backwards........who can blame these people for trying? And,  if we are talking about people from south america, it is our interference and OUR drug war that is making the area dangerous and poor and people dont want to live there! We shpould welcome them with open arms since we made a mess of their country!! I dont think we should do anything to some of these people. Especially if they have been here for awhile, certainly not prison!!!!  \\ 
  Topic 2 & Deport & I think they will probably be detained long enough awaiting trail and deportation and shouldn't serve any extra incarceration. I do not believe that process of trial and deportation would be instantaneous and I do not think that there needs to be and deterrent of extra jail time awarded if they are already going through the trial of being deported back to their home country. \\ 
  Topic 3 & Deport because of money & I am favor of just sending him back. Enough wasting tax payers money. Him living in USA prison is actually a higher standard of living than his country. He gets room and food everyday. \\ 
  Topic 4 & Depends on the circumstances & My first answer is no, but it also depends on why he illegally entered the U.S. If he committed a crime and fled to the U.S. then yes he should. If he came here for a better life, then I think that is something to be commended rather than punished. The people who would go that far to get better in life show hard work and dedication which America is supposed to be founded on. If I was a business owner, that is a man I would hire because he would strive for the best to keep his job because it meant a better life for him.  \\ 
  Topic 5 & More information needed & She did commit a crime but there could be a legitimate reason as to why she did so. She could be held until her background is checked and carefully monitored as to where bouts and work for so long and required to become a gainful citizen as everyone else. \\ 
  Topic 6 & Crime, small amount of jail time, then deportation & It doesn't seem as though the man poses a threat, so I'm reluctant to say that he deserves to be imprisoned. He did, however, enter the country illegally. When actual citizens break the law, they are sentenced to jail time, so I don't see why it should be any different with others. Also, if I were caught entering another country illegally, I would fully expect to face serious legal consequences. \\ 
  Topic 7 & Punish to full extent of the law & This person broke a law so that means they should be punished accordingly. Despite this person's history, this individual did something illegal and as with anyone else, they must serve the applicable sentence for the crime. \\ 
  Topic 8 & Allow to stay, no prison, rehabilitate, probably another explanation & We do not know what is her real situation. I have a friend graduated from one of the Ivy league schools, she taught in one universities in USA, her visa was expired just because she waited adjustment from Immigration, that means was not her fault at all, but at the end court called her, she had to be in court for several times before she decided to go home to her native country. Base on what she said, Immigration made tough access for skilled and educated people, they prefer illegal people with children. Therefore, government need to do something to fix this corrupt system. \\ 
  Topic 9 & No prison, deportation & he should be deported once again instead of being kept in prison and using our resources, it does not seem that he will be productive after another prison sentence \\ 
  Topic 10 & Should be sent back & I feel this person should be sent back to his own country.  I do not know of any punishment that would improve the situation.  If we imprison him in this country, we would have to accommodate him and pay for his food and essentials.  I feel that would cost far more than the cost of deporting him back to his country. \\ 
Topic 11 & Repeat offender, danger to society & This man appears to be disturbed in that he enters this country illegally and commits crimes while here.  I believe this person has a distorted view of how to live in this world and I do not think that he wants help nor does he want to live a law abiding life in the U.S.  He also, appears to be an obvious threat to others. Prison will probably not discourage this individual from entering illegally but a prison sentence might send a stronger message than simply being deported.  He did violate our laws when entering the country without permission.  This person's home country should step up and begin taking responsibility for their citizens and should try to monitor individuals deported back to the home country. \\ 
   \hline
\end{tabular}
\end{scriptsize}
\caption{Experiment 3: Topics and representative documents}
\label{topicsdocs3}
\end{table}


\end{document}